\documentclass[conference]{IEEEtran}
\IEEEoverridecommandlockouts

\usepackage{cite}
\usepackage{amsmath,amssymb,amsfonts}
\usepackage{algorithmic}
\usepackage{graphicx}
\usepackage{textcomp}
\usepackage{xcolor}
\usepackage{times}  \usepackage{helvet}  \usepackage{courier}  \usepackage[hyphens]{url}  \usepackage{graphicx}    \usepackage{caption} \usepackage{subcaption}

\usepackage{algorithm}
\usepackage{algorithmic}
\usepackage{booktabs}
\usepackage{adjustbox}
\usepackage{array}
\usepackage{amsmath} 

\usepackage{newfloat}
\usepackage{listings}

\newcommand{\mms}{\texttt{MMS}}

\newcommand{\omnilingual}{\texttt{Omnilingual}}
\newcommand{\canary}{\texttt{Canary}}

\newcommand{\saa}{\texttt{Speech Accent Archive}}

\newcommand{\ltwoarcticspon}{\texttt{L2-Arctic Spontaneous}}
\newcommand{\allstarlpp}{\texttt{ALLSTAR-LPP}}
\newcommand{\allstardhr}{\texttt{ALLSTAR-DHR}}
\newcommand{\allstarhtone}{\texttt{ALLSTAR-HT1}}
\newcommand{\allstarhttwo}{\texttt{ALLSTAR-HT2}}
\newcommand{\openslr}{\texttt{OpenSLR-83}}
\newcommand{\modeli}[2]{$m_{#1}=\text{#2}$}

\newcommand{\degradationvalue}[3]{\ensuremath{d(m_{#2}\to m_{#1}) = #3\%}}

\usepackage{eso-pic}
\newcommand{\IEEEcopyrightnotice}{\textcopyright\ 2026 IEEE. Personal use of this material is permitted. Permission from IEEE must be obtained for all other uses, in any current or future media, including reprinting/republishing this material for advertising or promotional purposes, creating new collective works, for resale or redistribution to servers or lists, or reuse of any copyrighted component of this work in other works.}

\begin{document}
\bstctlcite{IEEEexample:BSTcontrol}

\title{Connecting the Dots: Enduring Performance Gaps in ASR Systems Between \\ Standard and Minority English Varieties}

\author{
\begin{tabular}{c}
Alexander Metzger$^{1}$\IEEEauthorrefmark{1}\thanks{\IEEEauthorrefmark{1}These authors contributed equally.}, \quad Aruna Srivastava$^{1}$\IEEEauthorrefmark{1} \quad Ruslan Mukhamedvaleev$^{1}$ \quad Eunjung Yeo$^{2}$ \\[0.1em]
Syed Ishtiaque Ahmed$^{3}$ \quad Nina Markl$^{4}$ \quad Sachin Kumar$^{5}$ \quad Farhan Samir$^{3}$ \\[0.5em]
\normalsize
$^{1}$Koel Labs LLC \quad $^{2}$University of Texas at Austin \quad $^{3}$University of Toronto \\
\normalsize $^{4}$University of Essex \quad $^{5}$The Ohio State University \\[0.3em]
\end{tabular}
}

\maketitle
\AddToShipoutPictureFG*{\AtPageLowerLeft{\put(54,18){\parbox[b]{504bp}{\fontsize{7}{8}\selectfont\IEEEcopyrightnotice}}}}

\begin{abstract}
Automatic Speech Recognition (ASR) system evaluations have consistently revealed performance disparities along lines of race, gender, socioeconomic status, and language variety. While these disparities are well-documented, they are often (implicitly or explicitly) treated as issues that subsequent, larger, and more universal models will correct.  We challenge this assumption of steady progress in reducing accent-related performance disparities, conducting a longitudinal evaluation of 16 prominent systems released between 2021-2025 across 8 datasets representing diverse English accent varieties. We find that the performance disparity between Mainstream US English (MUSE) and minority accents is stubborn, not reliably declining as a function of time. Specifically, while systems are reliably performant on MUSE, minority accents are subject to risks of large performance degradations between generations of systems, sometimes worsening by 65\% absolute Word Error Rate. We argue that standard ASR benchmarks enact an implicit language policy privileging Mainstream US English varieties. As such, these performance disparities reflect systemic failures of language technology development infrastructure that disproportionately affect underrepresented speaker populations.\footnote{\url{https://github.com/smfsamir/longitudinal-asr-eval}} 

\end{abstract}

\begin{IEEEkeywords}
ASR, fairness, evaluation, sociolinguistics
\end{IEEEkeywords}

\section{Introduction} \label{sec:introduction}
Automatic Speech Recognition (ASR) systems are among one of the oldest, formative pursuits in computational linguistics research \cite{li2017divination}. Concerns about the limited robustness of these systems, owing to the relative homogeneity of their training datasets, have been prominent for decades \cite{benzeghiba_automatic_2007}. A wide array of studies have reported on performance disparities among important demographic constructs. At different points in time, English ASR systems have been shown to be worse for children and the elderly \cite{benzeghiba_automatic_2007,vipperla_longitudinal_2008}; for Black speakers compared to white speakers in the US \cite{koenecke_racial_2020,lai_exploring_2023}; for speakers from lower socioeconomic backgrounds \cite{cercas_curry_classist_2024}; and for speakers of stigmatized varieties of English \cite{markl_language_2022,graham_evaluating_2024}, among others. 
These studies are situated in particular temporal and technological contexts. As such, they may be interpreted as merely temporary shortcomings, or glitches \cite{benjamin2023race} that will be attenuated by future research and development.

Indeed by now there are a number of targeted interventions for resolving these shortcomings \cite{nakatumba2025systematic}, but there is no guarantee that these practices will be adopted in training the next generations of pretrained speech models. Instead, the more standard approach in training new generations of models is to train on larger datasets with larger models \cite{kaplan2020scaling}, speculating that their larger scale will also entail improved representation of minority accents. 
It is then natural to wonder whether the performance gap between hegemonic Mainstream US English and minority accent varieties may simply naturally dissipate over time. The optimism undergirding scaling heuristics is not entirely unfounded, optimizing large models on ever-larger big data archives has proven a useful heuristic when it comes to optimizing predictive performance on standard benchmarks \cite{recht2025actuary}. 

At the same time, it is unclear whether scaling heuristics would help in making predictions for inputs that are scarcely represented in standard evaluation harnesses, as we argue to be the case for minority accents in speech processing evaluations. That is, increased representation on some apparent axes (number of hours, number of languages, number of tasks), does not entail comparable improvements on other less-visible axes (minority accent representation). 
It then behooves us to investigate whether performance disparities are actually dissipating over time, as models grow larger \cite{puvvada_less_2024}, are trained on more tasks \cite{chu_qwen2-audio_2024} as well as datasets with a longer overall duration and more languages \cite{radford_robust_2022}. Taking a longitudinal view towards analyzing performance disparities is important, as it can inform researchers, developers, and policymakers as to whether the scaling modus operandi in machine-learning can be expected to naturally erode performance disparities until they are negligible, or whether we require more targeted interventions for obtaining minority accent representation.

In this work, we perform this much-needed longitudinal analysis, evaluating $16$ prominent models on datasets that are designed to capture a wide range of varieties of English. Crucially, we select datasets that represent both standard accents, specifically Mainstream US English and Standard Southern British English, or MUSE and SSBE respectively, as well as minority accents, enabling us to assess the performance gap. We show that the gap has endured over the course of these years, with even some of the latest generations of models exhibiting a gap of 10\% absolute WER, even for relatively simple scripted sentences, and even worse (20\%) for unscripted ones. We also observe that high quality performance on MUSE is durably prioritized as a feature in the development of new systems, with systems consistently achieving high-quality transcripts that are often close to 0\% WER rates. By comparison,  minority accent representation can deteriorate between generations of systems, sometimes by as much as 65\% of WER; as such, the performance gap not only endures but can increase with the release of newer generations of systems. Our findings call for greater transparency of speech processing systems that are deployed in downstream applications where speakers of marginalized English varieties would come into sensitive contact with them.

\section{ASR Evaluation Infrastructure} \label{sec:english-benchmarks}
To understand why accent-based disparities reflect systematic trends, rather than merely independent and episodic glitches, we will first consider current benchmarking practices in ASR development. ML benchmark datasets have been theorized as a form of \textit{infrastructure} \cite{10.1145/3442188.3445918,denton2020bringing}, in many senses of the concept \cite{bowker2000sorting}. Most importantly for our purposes, they undergird the development of new speech-processing models; and the human judgments that went into the data curation process  \cite{scheuerman2021datasets} become more invisible as they become naturalized as standard benchmark datasets \cite{denton2020bringing}. For ASR benchmarks, these are socio-linguistic judgments about the type of speech that is considered ``typical'' and thus important to model, and the types that can be made invisible and subsequently ignored. In effect, this process of formalization into a benchmark dataset can be interpreted as a type of language policy \cite{ASR_data_bias_markl_background}, specifying the spoken English varieties and contexts that are most appropriate. Many Global Englishes are foregone in the curation of standard ASR benchmarks.

As such, we argue that prominent evaluation benchmarks effectuate an Anglocentric language policy. While there are no firm prescriptions (e.g., by publishers, regulators or professional organisations) regarding benchmark selection, there can nevertheless be discernible norms within the field.
As observed in \cite{ASR_data_bias_markl_background}, ``the absence of an official policy...often serves only to reinforce the power and hegemony of prestige varieties, and marginalize others.'' We can thus analyze regularities in benchmark selection, and interpret these as a de facto language policy \cite{ASR_data_bias_markl_background}.

\subsection{Benchmark Selection}
\begin{table}[t]
\centering
\small
    \caption{Models evaluated in our diachronic analysis. Parentheses indicate additional versions from the same model family. The last column is the proportion of evaluation datasets used by the model authors for benchmarking that are explicitly concerned with multi-accent representation. The names of the datasets are included in Table \ref{tab:model-selection-details} in the Appendix.}
\begin{adjustbox}{max width=\linewidth}
\begin{tabular}{lclll}
        \toprule
        \textbf{Model} & \textbf{Release} & \textbf{Org.} & \textbf{Multi-Accent Evaluation} \\
        &    &  & \textbf{Datasets (ASR)} \\
        \midrule
          Deepspeech \cite{amodei_deep_2016} ($m_1$) & 2021-10 & Mozilla  & 2/6\\
         WavLM \cite{chen_wavlm_2021} ($m_2$)  & 2021-12 & Microsoft  & 0/1\\
         whisper \cite{radford_robust_2022} ($m_3, (m_4, m_8)$) & 2022-09 & OpenAI  & 4/14\\
         MMS \cite{pratap_scaling_2023} ($m_5$) & 2023-05 & Meta  & 0/5\\
         OWSM \cite{peng_reproducing_2023} ($m_6 (m_7, m_{10}, m_{12})$)& 2023-09 & CMU  & 1/8\\
         SeamlessM4T \cite{barrault2023seamless} ($m_9$) & 2023-12 &  Meta  & 0/1 \\
         Qwen-Audio \cite{chu_qwen2-audio_2024} ($m_{11} (m_{13})$) & 2024-08 & Alibaba & 1/3\\
         Canary-Qwen \cite{puvvada_less_2024} ($m_{14}$) &  2025-07 & NVIDIA & 3/9\\
         Chirp-3 \cite{zhang_google_2023} ($m_{15}$)& 2025-10 & Google  & 1/9\\ 
         Omnilingual \cite{omnilingual_asr_team_omnilingual_2025} ($m_{16}$) & 2025-11 & Meta  & 0/5\\
         \bottomrule
    \end{tabular}
    \end{adjustbox}
    \label{tab:model-selection}
\end{table}

Concretely, we study these regularities in benchmark usage by first selecting a range of $16$ prominent ASR models over the last 4 years, shown in Table~\ref{tab:model-selection}. 
We selected models from 2021 onwards, as this period is when large-scale pre-training emerged as the dominant paradigm in machine learning \cite{gururangan-etal-2020-dont}.\footnote{Though, the computational philosophy to take a data-centric approach to ASR has been around since the 1960s \cite{li2017divination}.} 
We selected organization that are notable for prominent speech-processing technologies: Mozilla, Microsoft, OpenAI, Meta, CMU, Alibaba, NVIDIA, and Google. 
We list the benchmarks used to evaluate the models’ English ASR capabilities, drawn from their respective technical reports, which are also cited in full detail in Table~\ref{tab:model-selection-details} in the Appendix. 
In the case of composite benchmarks like SUPERB \cite{superb_benchmark}, OpenASR \cite{openasr_benchmark}, and, SpeechStew \cite{chan2021speechstew}, we list all of the constituent datasets individually rather than the aggregated benchmark.
In Table \ref{tab:dataset-minority-representation-discourse} in the Appendix, we list the subset of benchmarks that are explicitly concerned with multi-accent representation.
The concern is not whether pluralistic benchmarks exist at all, but rather that they are peripheral -- they are sometimes used in evaluations of speech AI systems, and often not.

Consider that out of the $61$ model evaluations conducted across the technical reports enumerated in Table \ref{tab:model-selection}, only $12$ of them include some consideration of accent diversity. 
Of the $24$ distinct evaluation benchmark families (counting English-language subsets within their parent families), $7$ have pluricentric considerations of English varieties. 
Three of the systems (WavLM, MMS, Omnilingual) include no explicit considerations of accent pluralism. It almost goes without saying that Mainstream US English is never excluded from the evaluation infrastructure. 
Moreover, ASR performance on this variety is often assessed in multiple contexts, from spontaneous vs. read speech (e.g., Switchboard vs. Librispeech); to individual speakers vs. multiple (TedLIUM vs. CasualConversations). By comparison, if accent pluralism is considered in the evaluation, it tends to be in a more limited range of contexts and registers. 
For example, the OWSM and Qwen 2.5 models evaluates performance on CommonVoice, only assessing for read-speech of neutral-register Wikipedia sentences. 
Whisper \cite{radford_robust_2022} and Canary \cite{puvvada_less_2024}  diverge from this, evaluating on a variety of registers and contexts.

\subsection{Discussion}
Other benchmarks also sought to increase representational diversity, but along other axes than L2-English speaker representation. We discuss two of these major axes, number of hours and multilinguality, while noting that progress along them is separate from improvements in diverse accent representation.

\paragraph{Other representational axes --- multilinguality}  
One prominent axis has been multilinguality; for example, \textit{Common Voice}, \textit{Fleurs}.  
This is a reasonable, well-motivated axis to pursue, since a prominent critique of NLP as a discipline has been its focus on English datasets emerging from central Anglophone countries \cite{lucy-etal-2024-aboutme}, a critique that applies for audio datasets as well \cite{agnew2025sound}. 
In collecting datasets that contain a larger number of distinct languages, however, these works typically adopt the normative stance that languages are geographically and culturally homogenous artifacts \cite{schneider_methodological_2019}. 
Often this stance is implicit, but can be ascertained from the categorization of the speech data into different language bins, with no discussion of the particular varieties as well as limited sociodemographic information \cite{agnew2025sound}, e.g., in Multilingual Librispeech \cite{pratap2020mls}. 
Sometimes, this stance is explicit, for example the Fleurs dataset explicitly recruits ``native’’ speakers in constructing its multilingual speech evaluation datasets: ``We collected three recordings by three different native speakers'' \cite{conneau_fleurs_2023}. 
Inspection of the English speech in this dataset indicates a Mainstream US English variety. 
Thus, the construction of ``linguistic diversity’’ is complex and value-laden, and the representation of World Englishes may even be better in a monolingual dataset than one that contends to serve as evaluation infrastructure for ``universal’’ \cite{conneau_fleurs_2023} speech models. 

There are multilingual datasets that have more pluralistic representations of English, for example, CommonVoice \cite{commonvoice_datasets} and VoxPopuli \cite{wang-etal-2021-voxpopuli}. 
However, the most common accent representations in CommonVoice is by far Mainstream US English, based on audits of v7 and v13  \cite{markl-2022-mind,10.1145/3617694.3623258}. VoxPopuli contains 15 European accent varieties, but is a minority of the total transcribed corpus (29/1,800 hours). How these corpora are used matters. For example, MMS evaluates on CommonVoice over the whole corpus \cite{pratap_scaling_2023}. Since minority accent varieties are a small proportion of the 
English subset of the corpus, and necessarily an even smaller subset of the entire corpus, the average word error rate 
would be rendered ineffective in tracking robustness to accent variation, particularly variation outside 
of Mainstream US English. 
As recognized in \cite{birhane_values_2022}, operationalizing performance as ``correctness across individual predictions'' is a value-laden modus operandi, and reflects Anglocentric language policy due to the skewed constructions of these datasets.  
By comparison, the Whisper model presents their evaluation targeted on the English partitions of both VoxPopuli and Whisper, thus more effectively tracking accent robustness \cite{radford_robust_2022}. 
For this reason, we distinguish between CommonVoice  and CommonVoice-EN, depending on whether the subset was evaluated specifically or whether it was the entire set. We draw the same contrast for VoxPopuli and Voxpopuli-EN. 

\noindent \paragraph{Other representational axes --- total duration.} Other evaluation benchmarks have sought to increase diversity through increasing the total duration of the entire benchmark dataset, implementing the aphorism ``there is no data like more data'', attributed to Robert Mercer during his tenure at IBM's Continuous Speech Recognition team \cite{li2017divination},  as a data collection policy. The \textit{CALLHOME} \cite{linguistic_data_consortium_cabank_2013}, \textit{SwitchBoard} \cite{graff_switchboard-2_1998-1}, and WSJ \cite{paul_design_1992} corpora were funded by the US Department of Defense to advance speech recognition research from ``specific database inquiry
tasks, characterized by medium vocabularies'' \cite{paul_design_1992} towards open-ended, multi-speaker  dictation. These were characterized by increases in number of speakers, number of words in transcripts, and recording durations. For example, CALLHOME emphasized ``18.3 hours of transcribed spontaneous speech, comprising about 230,000 words'' \cite{canavan_callhome_1997}. In a similar vein, we have ``WSJ corpus will provide DARPA its first general-purpose English, large vocabulary,
natural language, high perplexity, corpus containing significant quantities of both speech data (400 hrs.) and text
data (47M words).'' The focus on quantification and scale as a stand-in for diversity continues to be pervasive in ML data collection \cite{scheuerman2021datasets,bender2021dangers}, like TedLium \cite{rousseau-etal-2012-ted}  and Librispeech \cite{librispeech_datasets}, both describing their datasets first and foremost in terms of scale  (``a total of 774
talks, representing 118 hours of speech''; ``contains 1000 hours of speech''), rather than other potentially salient characteristics, like speakers' demographic attributes. As previously observed by \cite{scheuerman2021datasets}, a foundational assumption in targeting large volumes of data is that such ``unconstrained'' data collection methods will naturally represent diversity on sociodemographic axes \cite{scheuerman2021datasets}. 
In these works, the disparities between the availabilities of different sociodemographic speaker groups is rarely acknowledged let alone interrogated. This inevitably  leads to the proliferation of speech benchmarks that have skewed representations of Mainstream English varieties \cite{markl-2022-mind}.

\paragraph{Implications of sporadic multi-accent representation.} Benchmarks serve a dual purpose in ML. Most commonly, they serve as evidence of 
improved performance, understood as the most common goal in ML \cite{birhane_values_2022}. 
But there are other commonplace values, including parameter and data efficiency, transparency, and theoretical analyzability \cite{birhane_values_2022}. 
In pursuing primary goals that are not standard benchmark dominance, benchmarks serve a role that is analogous to regression tests in software development -- preventing against performance degradations. 

Indeed, we observe benchmarks assuming this regression prevention role in some of these systems, that have primary goals other than achieving state of the art ASR performance. 
For example, Qwen-2 expands the set of speech-processing tasks that the model can complete \cite{chu_qwen2-audio_2024}. 
The OWSM system was meant to serve as an open-data, open-weights (near-)replication of OpenAI's Whisper \cite{peng_reproducing_2023}, the 
latter being trained on a dataset of $680,000$ hours of speech recordings  \cite{radford_robust_2022}. 
The MMS system was aimed at expanding the set of languages that ASR models can transcribe \cite{pratap_scaling_2023}. 
The models are often not the state-of-the art on some benchmarks, yet this is not understood as a failure, but rather evidence that the model did not regress too much along other performance axes of their primary one; for example, both the MMS and OWSM reports are transparent that Whisper outperforms them on some standard benchmarks. 
However, we see that performance degradations for standard varieties of English are thoroughly tested, while minority varieties are not. We thus test the following hypothesis (\textit{H1}): \textit{The performance gap between standard and minority varieties will \textit{not} be systematically diminished over time, as minority English varieties are more likely to be subjected to significant performance degradations in subsequent generations of systems.}

\section{Datasets}
\label{sec:datasets}

To select datasets for the evaluation, we compiled a list of multi-accent datasets, specifically those with speakers for whom English is a second language (L2-English speakers). We aim for datasets that are not considered standard benchmarks in the ASR community; none of the ones in our evaluation  were used to by the development teams for the systems listed in Table~\ref{tab:model-selection}. We built our evaluation corpora list through a combination of scholarly search engines and consulting with sociophoneticians. This aggregation phase left us with $39$ potential datasets. However, many of these were unsuitable for the purposes of our analysis. Several datasets lacked accent annotations for their recordings which is unsuitable for a comparative analysis between performance on standard and minority accents. In the same vein, other datasets did not have recordings from L1-speakers, precluding a comparative analysis. Other datasets required requests for access that are still pending. For the remaining datasets, we downloaded and listened to random samples, excluding ones that had an overrepresentation of noisy, unintelligible recordings. Ultimately, we used $4$ multi-accent corpora for our evaluation. 

Specifically, we use the Speech Accent Archive \cite[12 English varieties]{weinberger2015speech}; the L2-Arctic dataset \cite[6 varieties]{zhao2018l2}; the OpenSLR83 dataset \cite[6 varieties]{demirsahin-etal-2020-open}; and the ALLSTAR dataset \cite[22 varieties]{bradlow2010allstar}. L2-Arctic contains both scripted and unscripted portions, and we treat these as separate datasets for our analysis. Similarly, ALLSTAR is partitioned into 4 subsets, comprising recordings from speakers reading sentences from the novel The Little Prince (ALLSTAR-LPP); the Declaration of Human Rights (ALLSTAR-DHR); and two lists from the Hearing in Noise Test  \cite{nilsson1994development} (ALLSTAR-HT1; ALLSTAR-HT2). We treat these pre-existing partitions as separate datasets for our analysis. We describe these datasets further in our supplementary appendix.

 \section{Experiments and Results}
Here, we consider the series of models from Table~\ref{tab:model-selection}  ordered by their release date, with $m_1=\texttt{Deepspeech},\dots,m_n=\texttt{Omnilingual}$. We then compute model Word Error Rates (WERs) on every accent variety  recorded in every dataset we use for our evaluation. We then study the evolution of performance on minority accent varieties, with Mainstream US English (MUSE) as a control accent in each dataset. 

Our longitudinal study entails comparing models that were trained by different research teams, who are using different training and evaluation datasets and procedures.  
That is they were developed in heterogeneous situational contexts. 
Our aim is not to argue for particular training algorithms or dataset compositions over others, but rather a sociotechnical analysis 
characterize broader dynamics in the representation of minority accents. 
As such, comparing models developed in varying contexts is reasonable, especially considering that ML system development has long been fueled by a singular focus on ``single defined performance number'' \cite{donoho2023data}; in the case of speech recognition, the WER. 
The type of heterogeneous longitudinal comparison we are conducting has also been performed in prior work \cite{ott2022mapping}. 

\subsection{Longitudinal trends}
\newlength{\myfigheight}
\setlength{\myfigheight}{2in}
\begin{figure*}[]
    \begin{subfigure}[t]{0.48\textwidth}
        \centering
        \includegraphics[height=\myfigheight]{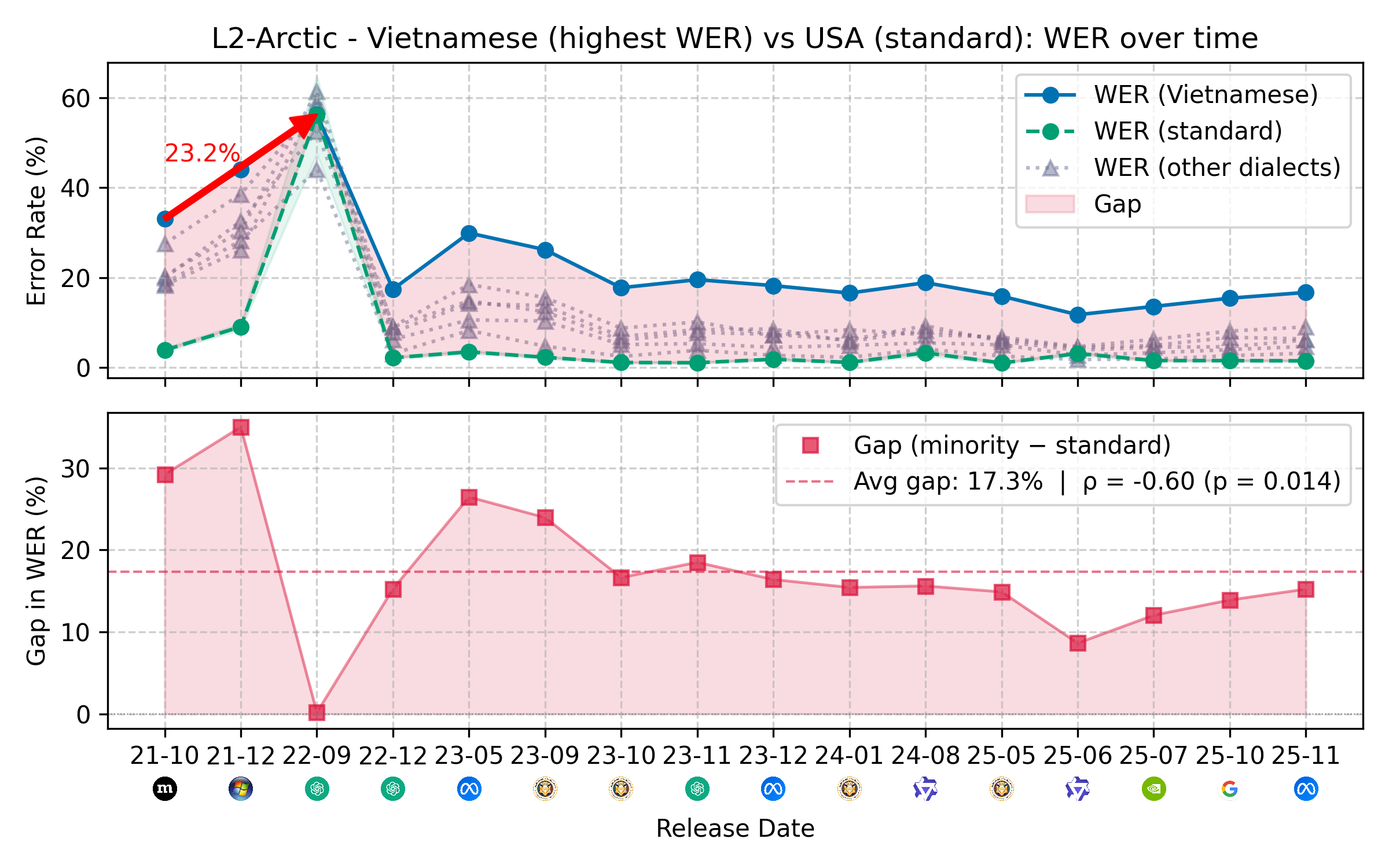}
    \end{subfigure}
    \hfill
    \begin{subfigure}[t]{0.48\textwidth}
    \centering
        \includegraphics[height=\myfigheight]{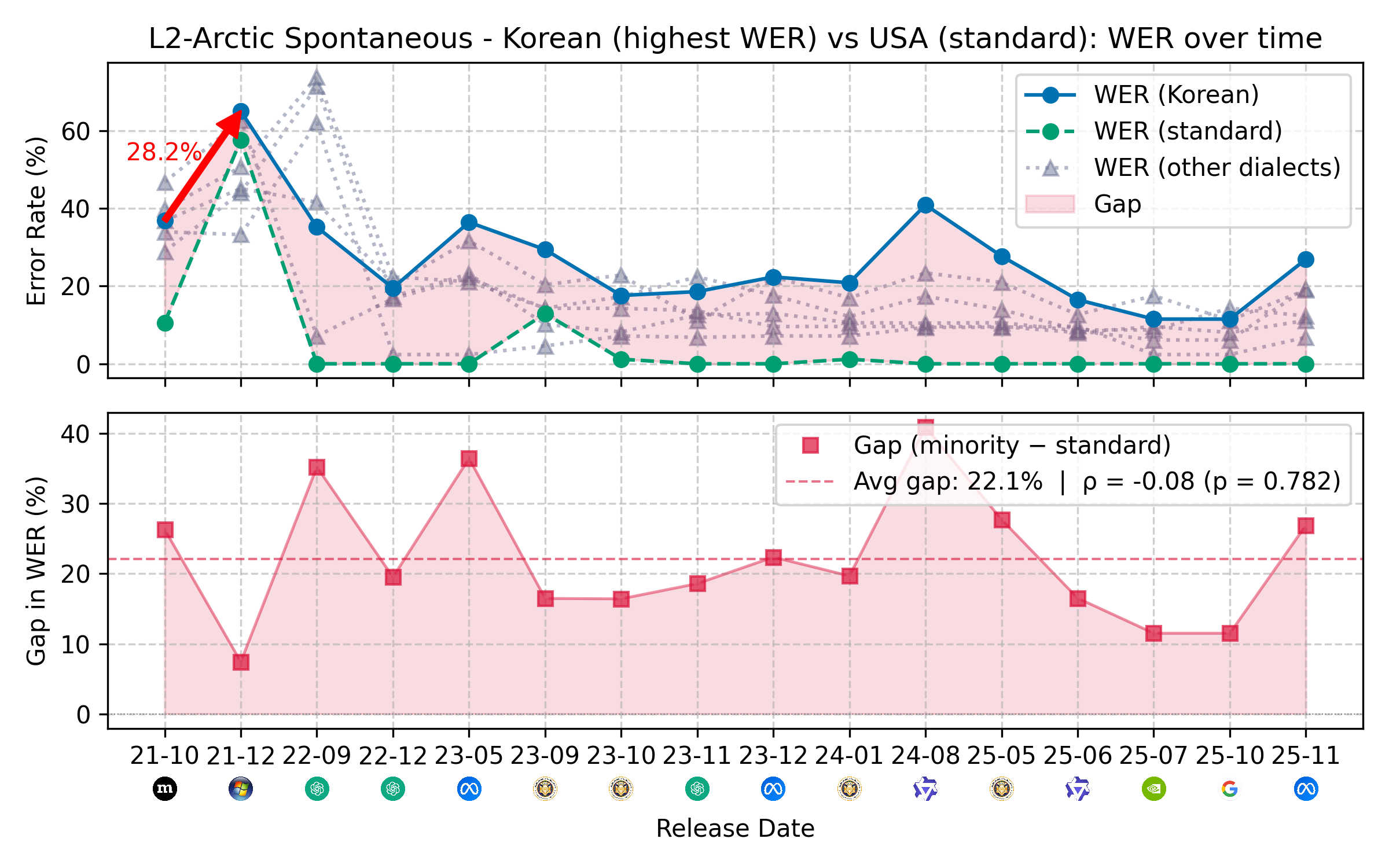}
    \end{subfigure}

    \begin{subfigure}[t]{0.48\textwidth}
        \centering
        \includegraphics[height=\myfigheight]{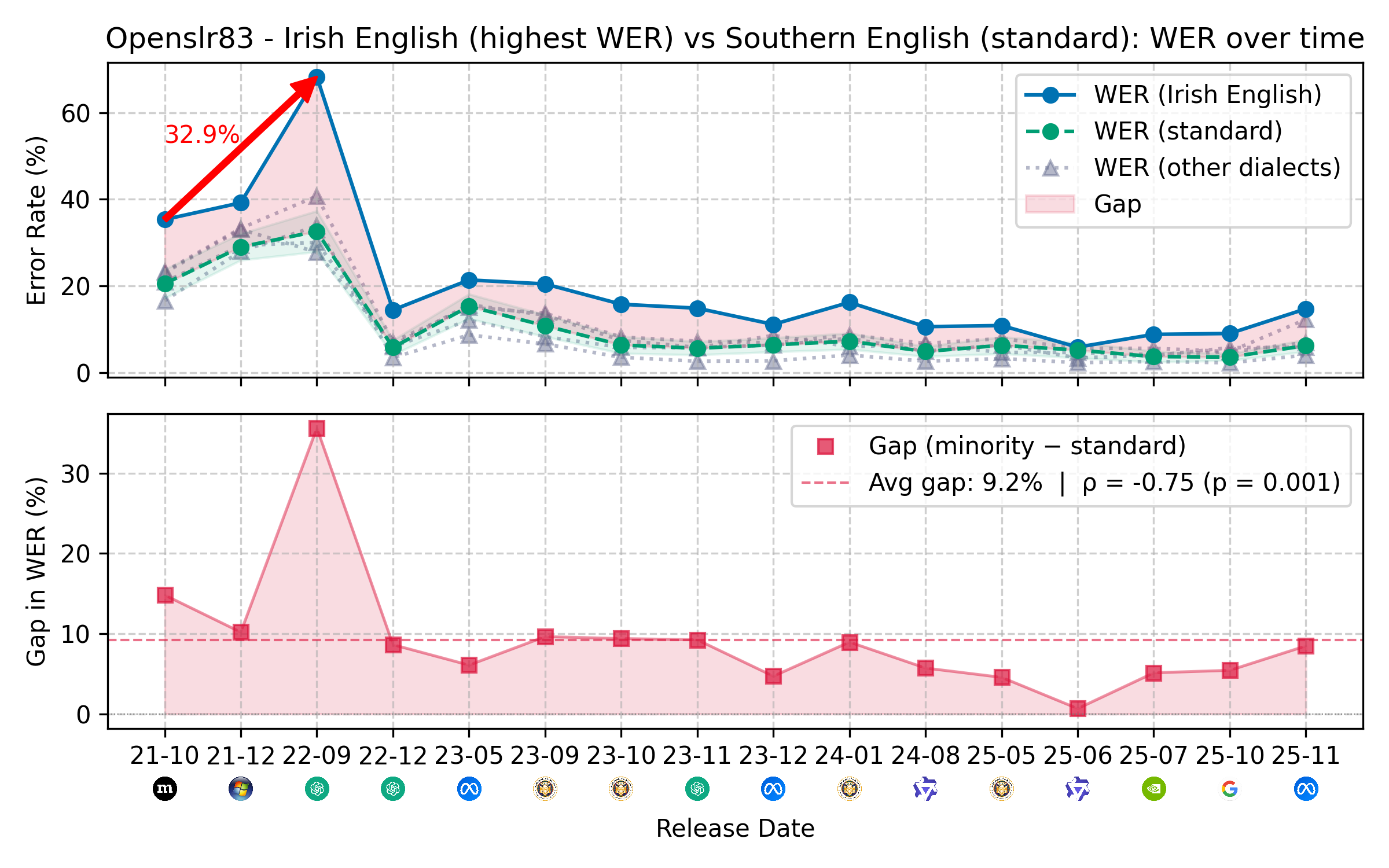}
    \end{subfigure}
    \hfill
    \begin{subfigure}[t]{0.48\textwidth}
        \centering
        \includegraphics[height=\myfigheight]{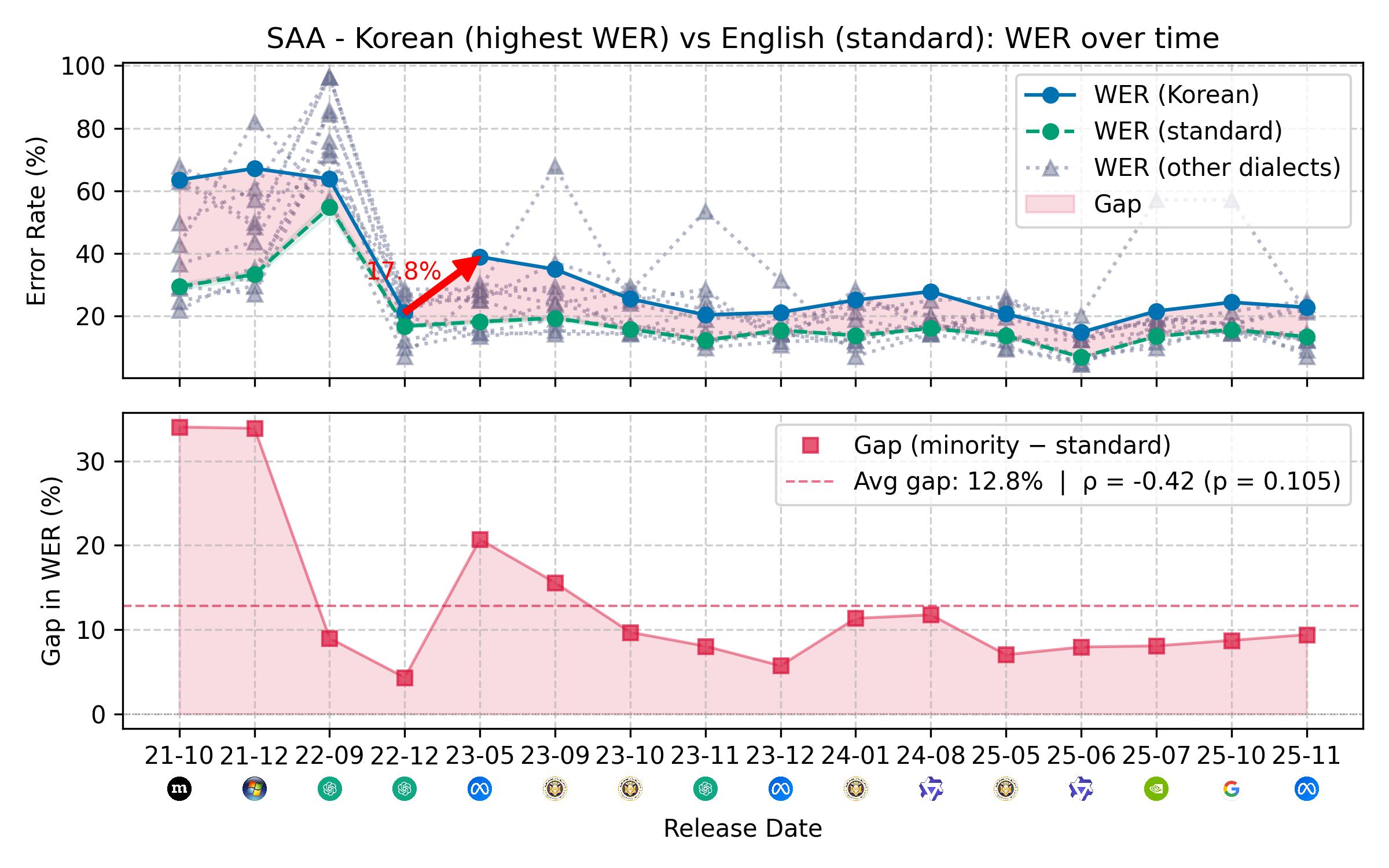}
    \end{subfigure}

    \begin{subfigure}[t]{0.48\textwidth}
        \centering
        \includegraphics[height=\myfigheight]{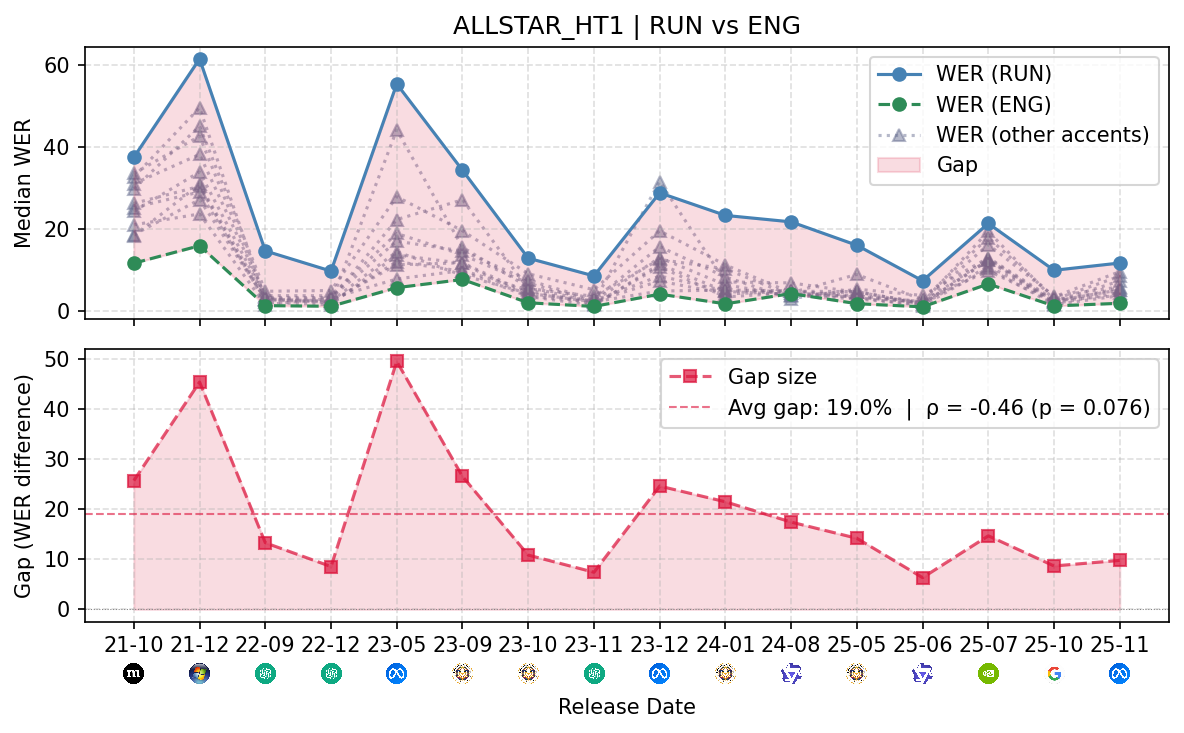}
    \end{subfigure}
    \hfill
    \begin{subfigure}[t]{0.48\textwidth}
        \centering
        \includegraphics[height=\myfigheight]{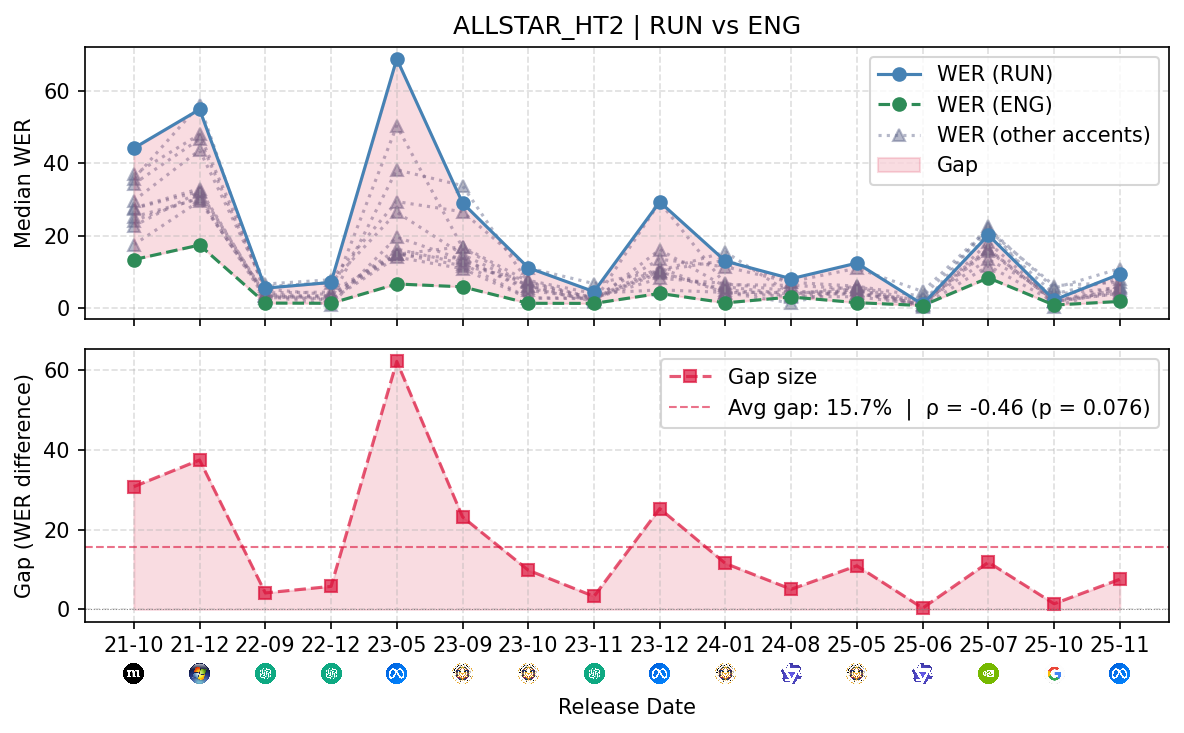}
    \end{subfigure}

    \begin{subfigure}[t]{0.48\textwidth}
        \centering
        \includegraphics[height=\myfigheight]{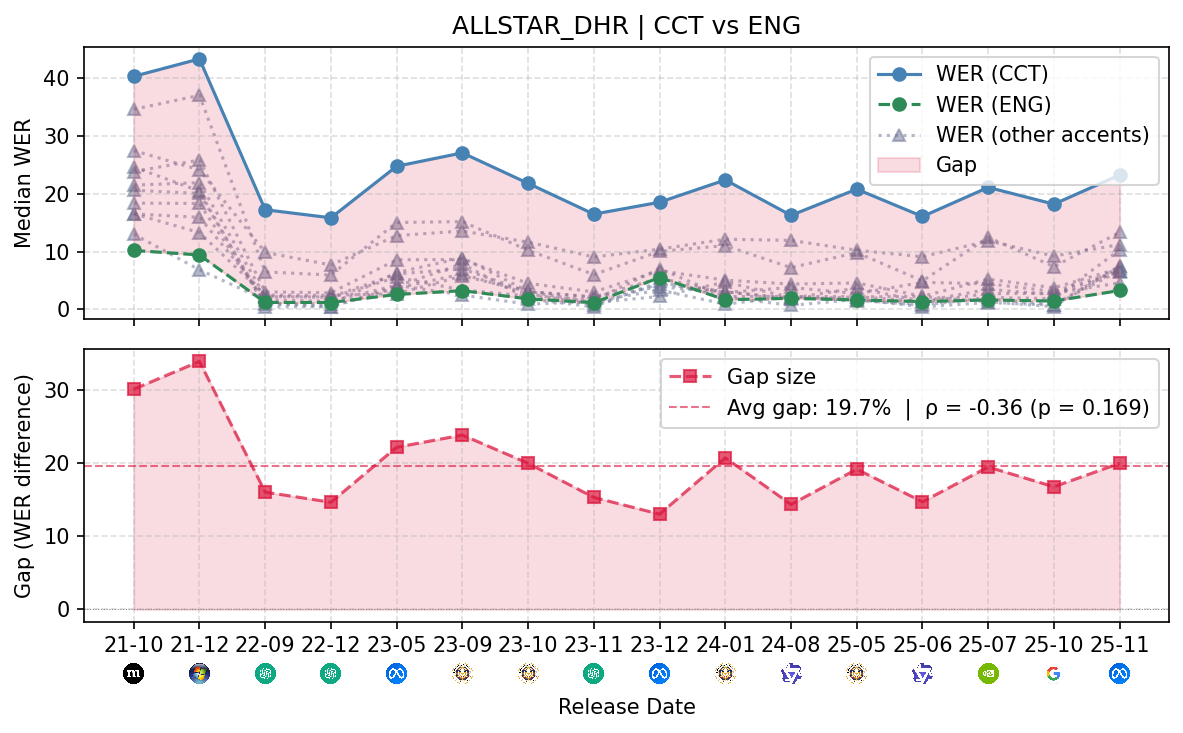}
    \end{subfigure}
    \hfill
    \begin{subfigure}[t]{0.48\textwidth}
        \centering
        \includegraphics[height=\myfigheight]{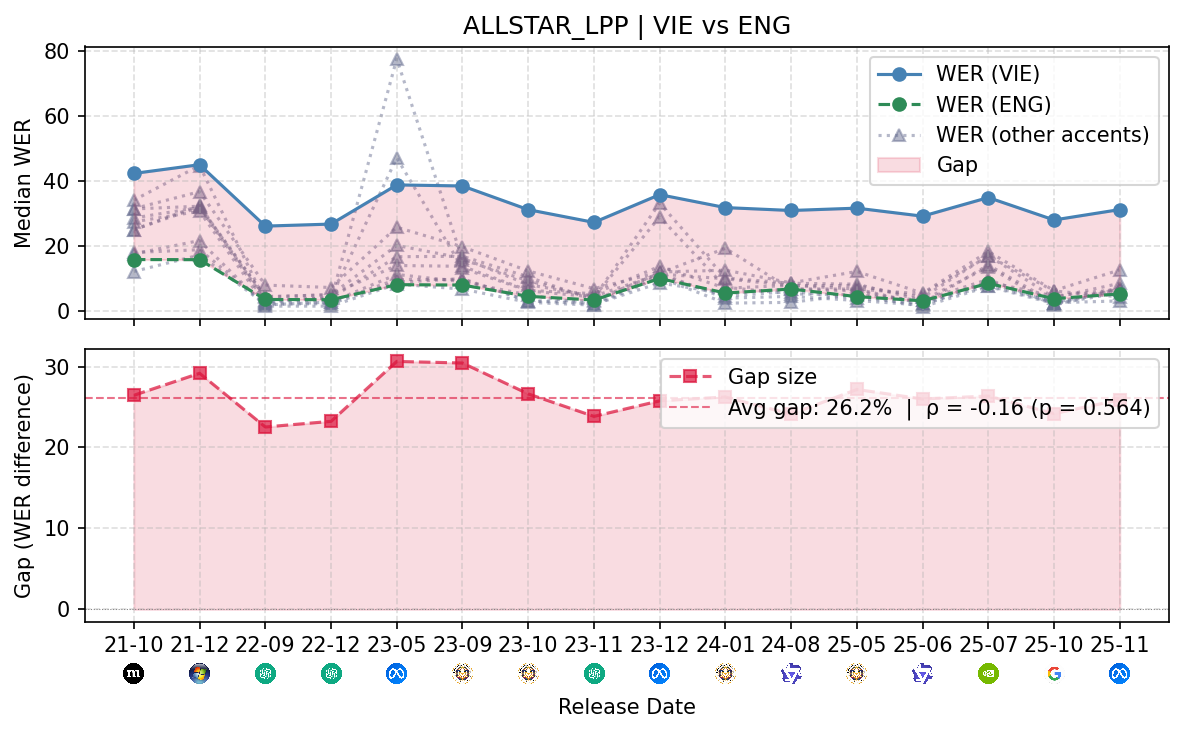}
    \end{subfigure}

    \vspace{-0.2em}
    \caption{
    Longitudinal system performances on 8 datasets. Top panels: Word Error Rates (WERs) for each accent variety. 
    Bottom panels: WER difference between Mainstream US English (MUSE) or Standard Southern British English (SSBE) and the minority accent
    variety with the highest average WER in the dataset. Bottom panels also show the average performance gap over the time period from the earliest evaluated model (2021-10) to the latest one (2025-11), as well as the correlation between the temporal ordering of the models and the size of the performance gap between the two accent varieties in each evaluation.
    }
    \label{fig:regression-plots}
\end{figure*}

We assess the extent to which the performance gap between 
standard accents (MUSE and SSBE) and minority accents has declined over the 
time period from $m_1$ to $m_n$. For a given dataset, we represent this performance gap as the 
WER difference between these standard accents and the accent variety with the 
highest average WER. We note that 
focusing on worst-case performance is a common practice in the ML fairness 
literature \cite{sagawa2019distributionally}; see also \cite{birhane_values_2022} for 
a critical discussion of the pitfalls of average-case metrics.

We highlight both of these accents for every dataset 
in Figure \ref{fig:regression-plots}. The top panel of each 
figure shows the WERs across each accent in the dataset, while 
the bottom panel shows the evolution of the performance gap 
between MUSE and the selected minority accent across 
$m_1,\dots,m_n$. We followed standard practice in ASR evaluation and applied a standard text normalization module prior to computing WER \cite{openasr_benchmark}.

\paragraph{Large average performance gaps across datasets.} As shown in Figure~\ref{fig:regression-plots}, we find that sizable performance gaps between standard accents and minority ones are present in every dataset that we evaluated across the time period represented by $m_1$ to $m_n$, with an average performance gap ranging from 9\% absolute WER, for SSBE compared to Irish English in the OpenSLR83 dataset, to $26\%$ for MUSE compared to Vietnamese English in the \textit{Little Prince} section of the ALLSTAR corpus.

\paragraph{Are the performance gaps declining over time?} To assess whether the performance gap is declining systematically over time, we computed the spearman correlation between ($\rho$) between the model-release dates and the standard-minority WER differences that constitute the performance gaps. We only find statistically significant evidence of the performance gap being ameliorated during this time period for two datasets, specifically for OpenSLR-83 and L2-Arctic with correlation values of $\rho=-0.75$ and $\rho=-0.60$ (both $p<0.05)$. For L2-Arctic, however, even the model ($m_{13}$) that performs the best at the selected minority accent, Vietnamese English, still exhibits a sizeable performance gap of $10\%$ WER. Meanwhile, for OpenSLR-83, while the performance gap has generally diminished over the full time period, the most recent releases we evaluated from $m_{13}$ to $m_{16}$ have bucked this trend, instead exacerbating the gap at each of these releases. In fact, the most recent evaluated model ($m_{16}$) exhibits a performance gap that matches the historical average performance gap of $9.2\%$ WER. For other datasets, evidence of the performance gap declining is not statistically significant. That is to say, there is overall scant evidence of the performance gap diminishing over time; instead of consistently declining, the trends in the performance gap size are highly non-monotonic.

\subsection{Large degradations across successive systems}
Here we look into the degradations that are the source of this non-monotonicity.
By degradation we mean a system at time $i$ that is worse 
than a system at time $j$, where $i>j$.
ML system development is a messy, loosely 
collaborative procedure, carried out by a large number of research teams in diverse settings \cite{donoho2023data}.
It is natural to expect some degradations in this noisy development process, especially since we
explicitly selected evaluation datasets that were not in the evaluation suites of the original development teams (see Table \ref{tab:model-selection}).
Even the MUSE variety has some degradations in these datasets, 
as can be observed for Whisper-V1 in the L2-Arctic evaluation (Figure 1; top-left panel).
But here, in alignment with our hypothesis (H1) we demonstrate experimentally that the regressions are 
much larger for minority accent varieties than for MUSE.
To do so we can consider the set of degradations for an accent variety:  
$$
\mathbb{D}=\{d(m_j\to m_i) \mid i\in[n],j\in[i], d(m_j\to m_i) > 0\}
$$
where $d(m_j \to m_i) = \text{WER}(m_i) - \text{WER}(m_j)$. We can then compute the average magnitude, or the \textit{Mean Degradation} (MD).

\begin{table}
    \centering
    \small
        \caption{Mean Degradation (MD) and MD rank.}
    \begin{tabular}{p{3cm}p{2cm}p{1cm}p{1cm}}
        \textbf{Dataset} & \textbf{Background} & $\mathbf{MD}$ & \textbf{Rank} \\
        \toprule
         L2-Arctic & Vietnamese En. & $\mathbf{11.4\%}$* & 5/7\\
         & MUSE  & $3.9\%$ & 1/7 \\
         \midrule
         L2-Arctic-Spontaneous & Korean En.  & $\mathbf{9.8\%}$* & 3/7 \\
          & MUSE  & $7.5\%$ & 2/7 \\
         \midrule
         Speech Accent Archive & Korean En. & $\mathbf{11.7\%}$ & 4/12 \\
        &  MUSE  & $11.6\%$ & 3/12\\
        \midrule 
        OpenSLR-83 & Irish En. & $\mathbf{16.5\%}$ & 6/6 \\
        & Southern En.  & $8.4\%$ & 5/6 \\
        \midrule
        ALLSTAR-LPP & Vietnamese En. & $\mathbf{3.6\%}$*  & 7/22 \\
        & MUSE  & $2.5\%$ & 2/22 \\
        \midrule
        ALLSTAR-DHR & Cantonese En. & $\mathbf{3.8\%}$*  & 17/21\\
        & MUSE & $1.2\%$ &  2/21 \\
        \midrule
        ALLSTAR-HT1 & Nkore En. & $\mathbf{11.8\%}$* & 22/22\\
         & MUSE & $2.3\%$ & 2/22\\
        \midrule
        ALLSTAR-HT2 & Nkore En. & $\mathbf{14.1\%}$* & 22/22\\ 
        & MUSE & $2.7\%$ & 2/22\\
        \bottomrule
    \end{tabular}
    \label{tab:degradation-table}
\end{table}

In Table \ref{tab:degradation-table}, we show the Mean Degradation for both the standard and the selected accent varieties. In all cases, we find the minority accent variety has a higher MD score, indicating performance degradations will be more impactful for these varieties. 
We test for statistical significance using a bootstrap 
percentile test \cite{efron1994introduction}: re-sampling 
the WER distributions for both varieties, 
re-computing their MD scores, and 
checking whether the minority accent variety 
has a higher MD than MUSE. 
We repeat this procedure with $B=1000$ bootstrap samples. 
We find that the procedure is statistically significant 
for all datasets except OpenSLR-83 and Speech Accent Archive (SAA). 
For OpenSLR-83, we find Irish English (minority) 
has similar regression patterns to 
Southern British English (standard). For SAA, 
the similar MD scores are explained by a large increase 
in error rates with $m_3=\text{\texttt{whisper-v1}}$ for MUSE. \texttt{whisper-v1} also performed poorly on 
Korean English, but since prior models ($m_1$, $m_2$) also 
struggled with Korean English, the regression 
($d(m_2 \to m_3)$ or $d(m_1 \to m_3)$) was not nearly as pronounced. 

\paragraph{Ordinal rankings of accent varieties.}
In each dataset, we rank the varieties in ascending order of MD.
MUSE ranks 1/7 for L2-Arctic, 2/7 for L2-Arctic Spontaneous,
3/12 for Speech Accent Archive, 2/22 for ALLSTAR-LPP,
2/21 for ALLSTAR-DHR, and 2/22 for both ALLSTAR-HT1 and ALLSTAR-HT2.
In OpenSLR-83, the Southern English control ranks 5/6 and Irish English ranks 6/6.
In every dataset, the selected minority accent variety has a higher MD rank
than its standard-accent control. These results support greater average
severity of performance degradations for the selected minority varieties.

\paragraph{Large degradations across successive systems.} 
The degradations between successive iterations of systems for minority accent varieties can be dramatic. 
We highlight the largest degradation for the selected minority accent variety  
in Figure~\ref{fig:regression-plots}. 
This is especially notable with \modeli{5}{\mms}, where \degradationvalue{5}{4}{18} (\saa), 
as well as  \degradationvalue{5}{4}{45} (\allstarhtone), 
\degradationvalue{5}{4}{65} (\allstarhttwo), \degradationvalue{5}{4}{7} (\allstarlpp), and 
\degradationvalue{5}{4}{9} (\allstardhr). 
We find that the \modeli{16}{\omnilingual} model, the most recent model we assessed, also degraded on some varieties: \degradationvalue{16}{14}{15.4} on \ltwoarcticspon; \degradationvalue{16}{15}{5.7} on \openslr. 
In all cases, the degradations for the MUSE variety are considerably smaller. 
\modeli{14}{\canary} is recognized as the 
top-performing multilingual ASR model on the OpenASR leaderboard, boasting a 5\% error rate on the OpenASR evaluation corpus. 
However, these error rates nearly double for Cantonese and Nkore-influenced accents in the ALLSTAR corpus 
(\degradationvalue{14}{3}{6.7} and \degradationvalue{14}{4}{10.1}).

\section{Conclusion}
A common fixture in the development process of novel ASR systems is evaluation on standard benchmarks. Among these standard benchmarks, Mainstream US English accents are always represented, often disproportionately so. This conventionalized cultural practice ensures that novel ASR systems exhibit stable  performance on this prestige variety. But other World Englishes are often marginalized in the computational operationalization of English, especially those in the Global Majority \cite{markl-2022-mind}. We study the effects of this systemic exclusion on a longitudinal basis, investigating performance on marginalized English accent varieties on a number of notable models across a critical four-year period (2021-2025), using sociolinguistically varied speech datasets that are outside of the standard benchmarking milieu. We find that while performance can be improved on these varieties, the selected minority accent varieties exhibit higher mean degradation than their standard-accent controls.

\bibliographystyle{IEEEtran}
\bibliography{mybib}

@IEEEtranBSTCTL{IEEEexample:BSTcontrol,
  CTLuse_forced_etal       = "yes",
  CTLmax_names_forced_etal = "3",
  CTLnames_show_etal       = "1"
}

@inproceedings{ASR_data_bias_markl_background,
    title = "Language technology practitioners as language managers: arbitrating data bias and predictive bias in {ASR}",
    author = "Markl, Nina  and
      McNulty, Stephen Joseph",
    editor = "Calzolari, Nicoletta  and
      B{\'e}chet, Fr{\'e}d{\'e}ric  and
      Blache, Philippe  and
      Choukri, Khalid  and
      Cieri, Christopher  and
      Declerck, Thierry  and
      Goggi, Sara  and
      Isahara, Hitoshi  and
      Maegaard, Bente  and
      Mariani, Joseph  and
      Mazo, H{\'e}l{\`e}ne  and
      Odijk, Jan  and
      Piperidis, Stelios",
    booktitle = "Proceedings of the Thirteenth Language Resources and Evaluation Conference",
    month = jun,
    year = "2022",
    address = "Marseille, France",
    publisher = "European Language Resources Association",
    url = "https://aclanthology.org/2022.lrec-1.680/",
    pages = "6328--6339",
}

@misc{openasr_benchmark,
      title={Open ASR Leaderboard: Towards Reproducible and Transparent Multilingual and Long-Form Speech Recognition Evaluation}, 
      author={Vaibhav Srivastav and Steven Zheng and Eric Bezzam and Eustache Le Bihan and Nithin Koluguri and Piotr Żelasko and Somshubra Majumdar and Adel Moumen and Sanchit Gandhi},
      year={2025},
      eprint={2510.06961},
      archivePrefix={arXiv},
      primaryClass={cs.CL},
      url={https://arxiv.org/abs/2510.06961}, 
}

@misc{superb_benchmark,
      title={SUPERB: Speech processing Universal PERformance Benchmark}, 
      author={Shu-wen Yang and Po-Han Chi and Yung-Sung Chuang and Cheng-I Jeff Lai and Kushal Lakhotia and Yist Y. Lin and Andy T. Liu and Jiatong Shi and Xuankai Chang and Guan-Ting Lin and Tzu-Hsien Huang and Wei-Cheng Tseng and Ko-tik Lee and Da-Rong Liu and Zili Huang and Shuyan Dong and Shang-Wen Li and Shinji Watanabe and Abdelrahman Mohamed and Hung-yi Lee},
      year={2021},
      eprint={2105.01051},
      archivePrefix={arXiv},
      primaryClass={cs.CL},
      url={https://arxiv.org/abs/2105.01051}, 
}

@inproceedings{librispeech_datasets,
	title = {Librispeech: {An} {ASR} corpus based on public domain audio books},
	shorttitle = {Librispeech},
	url = {https://ieeexplore.ieee.org/document/7178964},
	doi = {10.1109/ICASSP.2015.7178964},
	urldate = {2025-11-07},
	booktitle = {Proceedings of ICASSP},
	author = {Panayotov, Vassil and Chen, Guoguo and Povey, Daniel and Khudanpur, Sanjeev},
	month = apr,
	year = {2015},
	note = {ISSN: 2379-190X},
	pages = {5206--5210},
}

@inproceedings{commonvoice_datasets,
  author = {Ardila, R. and Branson, M. and Davis, K. and Henretty, M. and Kohler, M. and Meyer, J. and Morais, R. and Saunders, L. and Tyers, F. M. and Weber, G.},
  title = {Common Voice: A Massively-Multilingual Speech Corpus},
  booktitle = {Proceedings {LREC}},
  pages = {4211--4215},
  year = 2020
}

@article{kominek_cmu_database,
author = {Kominek, John and Black, Alan},
year = {2004},
month = {01},
pages = {},
title = {The CMU Arctic speech databases},
journal = {SSW5-2004}
}

@inproceedings{gigaspeech_dataset, 
    series={interspeech},
   title={GigaSpeech: An Evolving, Multi-Domain ASR Corpus with 10,000 Hours of Transcribed Audio},
   url={http://dx.doi.org/10.21437/Interspeech.2021-1965},
   DOI={10.21437/interspeech.2021-1965},
   booktitle={Interspeech},
   publisher={ISCA},
   author={Chen, Guoguo and Chai, Shuzhou and Wang, Guan-Bo and Du, Jiayu and Zhang, Wei-Qiang and Weng, Chao and Su, Dan and Povey, Daniel and Trmal, Jan and Zhang, Junbo and Jin, Mingjie and Khudanpur, Sanjeev and Watanabe, Shinji and Zhao, Shuaijiang and Zou, Wei and Li, Xiangang and Yao, Xuchen and Wang, Yongqing and You, Zhao and Yan, Zhiyong},
   year={2021},
   month=aug, collection={interspeech_2021} }

@article{derwing_suitcase_l1_l2, title={THE RELATIONSHIP BETWEEN L1 FLUENCY AND L2 FLUENCY DEVELOPMENT}, volume={31}, DOI={10.1017/S0272263109990015}, number={4}, journal={Studies in Second Language Acquisition}, author={Derwing, Tracey M. and Munro, Murray J. and Thomson, Ronald I. and Rossiter, Marian J.}, year={2009}, pages={533–557}}

@inproceedings{birhane_values_2022,
	address = {Seoul Republic of Korea},
	title = {The {Values} {Encoded} in {Machine} {Learning} {Research}},
	isbn = {978-1-4503-9352-2},
	url = {https://dl.acm.org/doi/10.1145/3531146.3533083},
	doi = {10.1145/3531146.3533083},
	language = {en},
	urldate = {2025-10-09},
	booktitle = {2022 {ACM} {Conference} on {Fairness} {Accountability} and {Transparency}},
	publisher = {ACM},
	author = {Birhane, Abeba and Kalluri, Pratyusha and Card, Dallas and Agnew, William and Dotan, Ravit and Bao, Michelle},
	month = jun,
	year = {2022},
	pages = {173--184},
}

@inproceedings{conneau_fleurs_2023,
	address = {Doha, Qatar},
	title = {{FLEURS}: {FEW}-{Shot} {Learning} {Evaluation} of {Universal} {Representations} of {Speech}},
	copyright = {https://doi.org/10.15223/policy-029},
	isbn = {9798350396904},
	shorttitle = {{FLEURS}},
	url = {https://ieeexplore.ieee.org/document/10023141/},
	doi = {10.1109/SLT54892.2023.10023141},
	urldate = {2025-10-09},
	booktitle = {Proceedings of {SLT}},
	author = {Conneau, Alexis and Ma, Min and Khanuja, Simran and Zhang, Yu and Axelrod, Vera and Dalmia, Siddharth and Riesa, Jason and Rivera, Clara and Bapna, Ankur},
	month = jan,
	year = {2023},
	pages = {798--805},
}

@Misc{voxforge_eval_dataset,
  author = {Voxforge.org},
  title = {Free Speech Recognition (Linux, Windows and Mac) - voxforge.org},
  howpublished = {\url{http://www.voxforge.org/}},
  note = {Accessed 1/1/2026}
}

@inproceedings{CHiME_eval_dataset,
  author={Jon Barker and Shinji Watanabe and Emmanuel Vincent and Jan Trmal},
  title={{The Fifth 'CHiME' Speech Separation and Recognition Challenge: Dataset, Task and Baselines}},
  year=2018,
  booktitle={Interspeech},
  pages={1561--1565},
  doi={10.21437/Interspeech.2018-1768}
}

@inproceedings{ami_meeting_eval_dataset,
author = {Carletta, Jean and Ashby, Simone and Bourban, Sebastien and Flynn, Mike and Guillemot, Mael and Hain, Thomas and Kadlec, Jaroslav and Karaiskos, Vasilis and Kraaij, Wessel and Kronenthal, Melissa and Lathoud, Guillaume and Lincoln, Mike and Lisowska, Agnes and McCowan, Iain and Post, Wilfried and Reidsma, Dennis and Wellner, Pierre},
title = {The AMI meeting corpus: a pre-announcement},
year = {2005},
isbn = {3540325492},
publisher = {Springer-Verlag},
address = {Berlin, Heidelberg},
url = {https://doi.org/10.1007/11677482_3},
doi = {10.1007/11677482_3},
}

@misc{earnings22_eval_dataset,
      title={Earnings-22: A Practical Benchmark for Accents in the Wild}, 
      author={Miguel Del Rio and Peter Ha and Quinten McNamara and Corey Miller and Shipra Chandra},
      year={2022},
      eprint={2203.15591},
      archivePrefix={arXiv},
      primaryClass={cs.CL},
      url={https://arxiv.org/abs/2203.15591}, 
}

@misc{musan_eval_dataset,
      title={MUSAN: A Music, Speech, and Noise Corpus}, 
      author={David Snyder and Guoguo Chen and Daniel Povey},
      year={2015},
      eprint={1510.08484},
      archivePrefix={arXiv},
      primaryClass={cs.SD},
      url={https://arxiv.org/abs/1510.08484}, 
}

@article{casual_conv_eval_dataset,
  title={Casual Conversations v2: Designing a large consent-driven dataset for AI fairness},
  author={Porgali, Bilal and Albiero, V{\'\i}tor and Stock, David and others},
  journal={arXiv preprint arXiv:2211.05809},
  year={2022}
}

@article{schneider_methodological_2019,
	title = {Methodological nationalism in {Linguistics}},
	volume = {76},
	issn = {03880001},
	url = {https://linkinghub.elsevier.com/retrieve/pii/S0388000117303480},
	doi = {10.1016/j.langsci.2018.05.006},
	language = {en},
	urldate = {2025-10-09},
	journal = {Language Sciences},
	author = {Schneider, Britta},
	month = nov,
	year = {2019},
	pages = {101169},
}

@article{graham_evaluating_2024,
	title = {Evaluating {OpenAI}'s {Whisper} {ASR}: {Performance} analysis across diverse accents and speaker traits},
	volume = {4},
	number = {2},
	journal = {JASA Express Letters},
	author = {Graham, Calbert and Roll, Nathan},
	year = {2024},
	note = {Publisher: AIP Publishing},
}

@misc{chu_qwen2-audio_2024,
	title = {Qwen2-{Audio} {Technical} {Report}},
	copyright = {Creative Commons Attribution 4.0 International},
	url = {https://arxiv.org/abs/2407.10759},
	doi = {10.48550/ARXIV.2407.10759},
	urldate = {2025-11-14},
	publisher = {arXiv},
	author = {Chu, Yunfei and Xu, Jin and Yang, Qian and Wei, Haojie and Wei, Xipin and Guo, Zhifang and Leng, Yichong and Lv, Yuanjun and He, Jinzheng and Lin, Junyang and Zhou, Chang and Zhou, Jingren},
	year = {2024},
	note = {Version Number: 1},
}

@inproceedings{amodei_deep_2016,
	title = {Deep speech 2: {End}-to-end speech recognition in english and mandarin},
	booktitle = {International conference on machine learning},
	publisher = {PMLR},
	author = {Amodei, Dario and Ananthanarayanan, Sundaram and Anubhai, Rishita and Bai, Jingliang and Battenberg, Eric and Case, Carl and Casper, Jared and Catanzaro, Bryan and Cheng, Qiang and Chen, Guoliang and {others}},
	year = {2016},
	pages = {173--182},
}

@article{chen_wavlm_2021,
	title = {{WavLM}: {Large}-{Scale} {Self}-{Supervised} {Pre}-{Training} for {Full} {Stack} {Speech} {Processing}},
	copyright = {arXiv.org perpetual, non-exclusive license},
	shorttitle = {{WavLM}},
	url = {https://arxiv.org/abs/2110.13900},
	doi = {10.48550/ARXIV.2110.13900},
	urldate = {2025-12-22},
	author = {Chen, Sanyuan and Wang, Chengyi and Chen, Zhengyang and Wu, Yu and Liu, Shujie and Chen, Zhuo and Li, Jinyu and Kanda, Naoyuki and Yoshioka, Takuya and Xiao, Xiong and Wu, Jian and Zhou, Long and Ren, Shuo and Qian, Yanmin and Qian, Yao and Zeng, Michael and Yu, Xiangzhan and Wei, Furu},
	year = {2021},
	note = {Publisher: arXiv
Version Number: 5},
}

@misc{pratap_scaling_2023,
	title = {Scaling {Speech} {Technology} to 1,000+ {Languages}},
	copyright = {arXiv.org perpetual, non-exclusive license},
	url = {https://arxiv.org/abs/2305.13516},
	doi = {10.48550/ARXIV.2305.13516},
	urldate = {2025-12-22},
	publisher = {arXiv},
	author = {Pratap, Vineel and Tjandra, Andros and Shi, Bowen and Tomasello, Paden and Babu, Arun and Kundu, Sayani and Elkahky, Ali and Ni, Zhaoheng and Vyas, Apoorv and Fazel-Zarandi, Maryam and Baevski, Alexei and Adi, Yossi and Zhang, Xiaohui and Hsu, Wei-Ning and Conneau, Alexis and Auli, Michael},
	year = {2023},
	note = {Version Number: 1},
}

@misc{radford_robust_2022,
	title = {Robust {Speech} {Recognition} via {Large}-{Scale} {Weak} {Supervision}},
	copyright = {arXiv.org perpetual, non-exclusive license},
	url = {https://arxiv.org/abs/2212.04356},
	doi = {10.48550/ARXIV.2212.04356},
	urldate = {2025-12-22},
	publisher = {arXiv},
	author = {Radford, Alec and Kim, Jong Wook and Xu, Tao and Brockman, Greg and McLeavey, Christine and Sutskever, Ilya},
	year = {2022},
	note = {Version Number: 1},
}

@misc{peng_reproducing_2023,
	title = {Reproducing {Whisper}-{Style} {Training} {Using} an {Open}-{Source} {Toolkit} and {Publicly} {Available} {Data}},
	copyright = {arXiv.org perpetual, non-exclusive license},
	url = {https://arxiv.org/abs/2309.13876},
	doi = {10.48550/ARXIV.2309.13876},
	urldate = {2025-12-22},
	publisher = {arXiv},
	author = {Peng, Yifan and Tian, Jinchuan and Yan, Brian and Berrebbi, Dan and Chang, Xuankai and Li, Xinjian and Shi, Jiatong and Arora, Siddhant and Chen, William and Sharma, Roshan and Zhang, Wangyou and Sudo, Yui and Shakeel, Muhammad and Jung, Jee-weon and Maiti, Soumi and Watanabe, Shinji},
	year = {2023},
	note = {Version Number: 3},
}

@misc{bradlow2010allstar,
  author       = {Bradlow, A. R.},
  title        = {ALLSSTAR: Archive of L1 and L2 Scripted and Spontaneous Transcripts And Recordings},
  howpublished = {\url{https://speechbox.linguistics.northwestern.edu/allsstar}},
  note         = {Accessed: 2025.}, 
  year         = {n.d.}
}

@misc{zhang_google_2023,
	title = {Google {USM}: {Scaling} {Automatic} {Speech} {Recognition} {Beyond} 100 {Languages}},
	copyright = {arXiv.org perpetual, non-exclusive license},
	shorttitle = {Google {USM}},
	url = {https://arxiv.org/abs/2303.01037},
	doi = {10.48550/ARXIV.2303.01037},
	urldate = {2025-12-22},
	publisher = {arXiv},
	author = {Zhang, Yu and Han, Wei and Qin, James and Wang, Yongqiang and Bapna, Ankur and Chen, Zhehuai and Chen, Nanxin and Li, Bo and Axelrod, Vera and Wang, Gary and Meng, Zhong and Hu, Ke and Rosenberg, Andrew and Prabhavalkar, Rohit and Park, Daniel S. and Haghani, Parisa and Riesa, Jason and Perng, Ginger and Soltau, Hagen and Strohman, Trevor and Ramabhadran, Bhuvana and Sainath, Tara and Moreno, Pedro and Chiu, Chung-Cheng and Schalkwyk, Johan and Beaufays, Françoise and Wu, Yonghui},
	year = {2023},
	note = {Version Number: 3},
}

@misc{puvvada_less_2024,
	title = {Less is {More}: {Accurate} {Speech} {Recognition} \&amp; {Translation} without {Web}-{Scale} {Data}},
	copyright = {Creative Commons Attribution 4.0 International},
	shorttitle = {Less is {More}},
	url = {https://arxiv.org/abs/2406.19674},
	doi = {10.48550/ARXIV.2406.19674},
	urldate = {2025-12-22},
	publisher = {arXiv},
	author = {Puvvada, Krishna C. and Żelasko, Piotr and Huang, He and Hrinchuk, Oleksii and Koluguri, Nithin Rao and Dhawan, Kunal and Majumdar, Somshubra and Rastorgueva, Elena and Chen, Zhehuai and Lavrukhin, Vitaly and Balam, Jagadeesh and Ginsburg, Boris},
	year = {2024},
	note = {Version Number: 1},
}

@misc{omnilingual_asr_team_omnilingual_2025,
	title = {Omnilingual {ASR}: {Open}-{Source} {Multilingual} {Speech} {Recognition} for 1600+ {Languages}},
	copyright = {Creative Commons Attribution Share Alike 4.0 International},
	shorttitle = {Omnilingual {ASR}},
	url = {https://arxiv.org/abs/2511.09690},
	doi = {10.48550/ARXIV.2511.09690},
	urldate = {2025-12-22},
	publisher = {arXiv},
	author = {{Omnilingual ASR team} and Keren, Gil and Kozhevnikov, Artyom and Meng, Yen and Ropers, Christophe and Setzler, Matthew and Wang, Skyler and Adebara, Ife and Auli, Michael and Balioglu, Can and Chan, Kevin and Cheng, Chierh and Chuang, Joe and Droof, Caley and Duppenthaler, Mark and Duquenne, Paul-Ambroise and Erben, Alexander and Gao, Cynthia and Gonzalez, Gabriel Mejia and Lyu, Kehan and Miglani, Sagar and Pratap, Vineel and Sadagopan, Kaushik Ram and Saleem, Safiyyah and Turkatenko, Arina and Ventayol-Boada, Albert and Yong, Zheng-Xin and Chung, Yu-An and Maillard, Jean and Moritz, Rashel and Mourachko, Alexandre and Williamson, Mary and Yates, Shireen},
	year = {2025},
	note = {Version Number: 1},
}

@phdthesis{li2017divination,
  title={Divination engines: A media history of text prediction},
  author={Li, Xiaochang},
  year={2017},
  school={New York University}
}

@inproceedings{cercas_curry_classist_2024,
	address = {Bangkok, Thailand},
	title = {Classist {Tools}: {Social} {Class} {Correlates} with {Performance} in {NLP}},
	url = {https://aclanthology.org/2024.acl-long.682/},
	doi = {10.18653/v1/2024.acl-long.682},
	booktitle = {Proceedings of the 62nd {Annual} {Meeting} of the {Association} for {Computational} {Linguistics} ({Volume} 1: {Long} {Papers})},
	publisher = {Association for Computational Linguistics},
	author = {Cercas Curry, Amanda and Attanasio, Giuseppe and Talat, Zeerak and Hovy, Dirk},
	editor = {Ku, Lun-Wei and Martins, Andre and Srikumar, Vivek},
	month = aug,
	year = {2024},
	pages = {12643--12655},
}

@article{koenecke_racial_2020,
	title = {Racial disparities in automated speech recognition},
	volume = {117},
	issn = {0027-8424, 1091-6490},
	url = {https://pnas.org/doi/full/10.1073/pnas.1915768117},
	doi = {10.1073/pnas.1915768117},
	language = {en},
	number = {14},
	urldate = {2025-12-22},
	journal = {Proceedings of the National Academy of Sciences},
	author = {Koenecke, Allison and Nam, Andrew and Lake, Emily and Nudell, Joe and Quartey, Minnie and Mengesha, Zion and Toups, Connor and Rickford, John R. and Jurafsky, Dan and Goel, Sharad},
	month = apr,
	year = {2020},
	pages = {7684--7689},
}

@inproceedings{markl_language_2022,
	title = {Language variation and algorithmic bias: understanding algorithmic bias in {British} {English} automatic speech recognition},
	booktitle = {Proceedings of the 2022 {ACM} {Conference} on {Fairness}, {Accountability}, and {Transparency}},
	author = {Markl, Nina},
	year = {2022},
	pages = {521--534},
}

@inproceedings{lai_exploring_2023,
	title = {Exploring {Sources} of {Racial} {Bias} in {Automatic} {Speech} {Recognition} through the {Lens} of {Rhythmic} {Variation}.},
	volume = {2023},
	booktitle = {{Interspeech}},
	author = {Lai, Li-Fang and Holliday, Nicole R},
	year = {2023},
	pages = {1284--1288},
}

@article{benzeghiba_automatic_2007,
	title = {Automatic speech recognition and speech variability: {A} review},
	volume = {49},
	number = {10-11},
	journal = {Speech communication},
	author = {Benzeghiba, Mohamed and De Mori, Renato and Deroo, Olivier and Dupont, Stephane and Erbes, Teodora and Jouvet, Denis and Fissore, Luciano and Laface, Pietro and Mertins, Alfred and Ris, Christophe and {others}},
	year = {2007},
	note = {Publisher: Elsevier},
	pages = {763--786},
}

@inproceedings{vipperla_longitudinal_2008,
title     = {Longitudinal study of {ASR} performance on ageing voices},
  author    = {Ravichander Vipperla and Steve Renals and Joe Frankel},
  year      = {2008},
  booktitle = {Interspeech},
  pages     = {2550--2553},
  doi       = {10.21437/Interspeech.2008-632},
  issn      = {2958-1796},
}

@inproceedings{10.1145/3442188.3445918,
author = {Hutchinson, Ben and Smart, Andrew and Hanna, Alex and Denton, Remi and Greer, Christina and Kjartansson, Oddur and Barnes, Parker and Mitchell, Margaret},
title = {Towards Accountability for Machine Learning Datasets: Practices from Software Engineering and Infrastructure},
year = {2021},
isbn = {9781450383097},
publisher = {Association for Computing Machinery},
address = {New York, NY, USA},
url = {https://doi.org/10.1145/3442188.3445918},
doi = {10.1145/3442188.3445918},
booktitle = {Proceedings of the 2021 ACM Conference on Fairness, Accountability, and Transparency},
pages = {560–575},
numpages = {16},
location = {Virtual Event, Canada},
series = {FAccT '21}
}

@article{denton2020bringing,
  title={Bringing the people back in: Contesting benchmark machine learning datasets},
  author={Denton, Remi and Hanna, Alex and Amironesei, Razvan and Smart, Andrew and Nicole, Hilary and Scheuerman, Morgan Klaus},
  journal={arXiv preprint arXiv:2007.07399},
  year={2020}
}

@book{bowker2000sorting,
  title={Sorting things out: Classification and its consequences},
  author={Bowker, Geoffrey C and Star, Susan Leigh},
  year={2000},
  publisher={MIT press}
}

@article{donoho2023data,
  title={Data science at the singularity},
  author={Donoho, David},
  journal={arXiv preprint arXiv:2310.00865},
  year={2023}
}

@article{scheuerman2021datasets,
  title={Do datasets have politics? Disciplinary values in computer vision dataset development},
  author={Scheuerman, Morgan Klaus and Hanna, Alex and Denton, Remi},
  journal={Proceedings of the ACM on Human-Computer Interaction},
  volume={5},
  number={CSCW2},
  pages={1--37},
  year={2021},
  publisher={ACM New York, NY, USA}
}

@inproceedings{markl-2022-mind,
    title = "Mind the data gap(s): Investigating power in speech and language datasets",
    author = "Markl, Nina",
    editor = "Chakravarthi, Bharathi Raja  and
      Bharathi, B  and
      McCrae, John P  and
      Zarrouk, Manel  and
      Bali, Kalika  and
      Buitelaar, Paul",
    booktitle = "Proceedings of the Second Workshop on Language Technology for Equality, Diversity and Inclusion",
    month = may,
    year = "2022",
    address = "Dublin, Ireland",
    publisher = "Association for Computational Linguistics",
    url = "https://aclanthology.org/2022.ltedi-1.1/",
    doi = "10.18653/v1/2022.ltedi-1.1",
    pages = "1--12"
}

@article{dichristofano2022global,
  title={Global performance disparities between english-language accents in automatic speech recognition},
  author={DiChristofano, Alex and Shuster, Henry and Chandra, Shefali and Patwari, Neal},
  journal={arXiv preprint arXiv:2208.01157},
  year={2022}
}

@inproceedings{chan2022training,
  title={Training and typological bias in ASR performance for world Englishes.},
  author={Chan, May Pik Yu and Choe, June and Li, Aini and Chen, Yiran and Gao, Xin and Holliday, Nicole R},
  booktitle={Interspeech},
  pages={1273--1277},
  year={2022}
}

@article{weinberger2015speech,
  title={Speech accent archive},
  author={Weinberger, Steven},
  journal={George Mason University},
  year={2015}
}

@article{nakatumba2025systematic,
  title={A Systematic Literature Review on Bias Evaluation and Mitigation in Automatic Speech Recognition Models for Low-Resource African Languages},
  author={Nakatumba-Nabende, Joyce and Kagumire, Sulaiman and Kantono, Caroline and Nabende, Peter},
  journal={ACM Computing Surveys},
  volume={58},
  number={4},
  pages={1--24},
  year={2025},
  publisher={ACM New York, NY}
}

@incollection{benjamin2023race,
  title={Race after technology},
  author={Benjamin, Ruha},
  booktitle={Social Theory Re-Wired},
  pages={405--415},
  year={2023},
  publisher={Routledge}
}

@inproceedings{agnew2025sound,
  title={Sound Check: Auditing Recent Audio Dataset Practices},
  author={Agnew, William and Barnett, Julia and Chu, Annie and Hong, Rachel and Feffer, Michael and Netzorg, Robin and Jiang, Harry H and Awumey, Ezra and Das, Sauvik},
  booktitle={Proceedings of the AAAI/ACM Conference on AI, Ethics, and Society},
  volume={8},
  number={1},
  pages={26--40},
  year={2025}
}

@article{pratap2020mls,
  title={Mls: A large-scale multilingual dataset for speech research},
  author={Pratap, Vineel and Xu, Qiantong and Sriram, Anuroop and Synnaeve, Gabriel and Collobert, Ronan},
  journal={arXiv preprint arXiv:2012.03411},
  year={2020}
}

@inproceedings{wang-etal-2021-voxpopuli,
    title = "{V}ox{P}opuli: A Large-Scale Multilingual Speech Corpus for Representation Learning, Semi-Supervised Learning and Interpretation",
    author = "Wang, Changhan  and
      Riviere, Morgane  and
      Lee, Ann  and
      Wu, Anne  and
      Talnikar, Chaitanya  and
      Haziza, Daniel  and
      Williamson, Mary  and
      Pino, Juan  and
      Dupoux, Emmanuel",
    editor = "Zong, Chengqing  and
      Xia, Fei  and
      Li, Wenjie  and
      Navigli, Roberto",
    booktitle = "Proceedings of the 59th Annual Meeting of the Association for Computational Linguistics and the 11th International Joint Conference on Natural Language Processing (Volume 1: Long Papers)",
    month = aug,
    year = "2021",
    address = "Online",
    publisher = "Association for Computational Linguistics",
    url = "https://aclanthology.org/2021.acl-long.80/",
    doi = "10.18653/v1/2021.acl-long.80",
    pages = "993--1003"
}

@inproceedings{10.1145/3617694.3623258,
author = {Reid, Kathy and Williams, Elizabeth T.},
title = {Common Voice and accent choice: data contributors self-describe their spoken accents in diverse ways},
year = {2023},
isbn = {9798400703812},
publisher = {Association for Computing Machinery},
address = {New York, NY, USA},
url = {https://doi.org/10.1145/3617694.3623258},
doi = {10.1145/3617694.3623258},
booktitle = {Proceedings of the 3rd ACM Conference on Equity and Access in Algorithms, Mechanisms, and Optimization},
articleno = {35},
numpages = {10},
location = {Boston, MA, USA},
series = {EAAMO '23}
}

@inproceedings{lucy-etal-2024-aboutme,
    title = "{A}bout{M}e: Using Self-Descriptions in Webpages to Document the Effects of {E}nglish Pretraining Data Filters",
    author = "Lucy, Li  and
      Gururangan, Suchin  and
      Soldaini, Luca  and
      Strubell, Emma  and
      Bamman, David  and
      Klein, Lauren  and
      Dodge, Jesse",
    editor = "Ku, Lun-Wei  and
      Martins, Andre  and
      Srikumar, Vivek",
    booktitle = "Proceedings of the 62nd Annual Meeting of the Association for Computational Linguistics (Volume 1: Long Papers)",
    month = aug,
    year = "2024",
    address = "Bangkok, Thailand",
    publisher = "Association for Computational Linguistics",
    url = "https://aclanthology.org/2024.acl-long.400/",
    doi = "10.18653/v1/2024.acl-long.400",
    pages = "7393--7420"
}

@misc{linguistic_data_consortium_cabank_2013,
	title = {{CABank} {English} {CallHome} {Corpus}},
	url = {https://ca.talkbank.org/access/CallHome/eng.html},
	doi = {10.21415/T5KP54},
	urldate = {2026-01-03},
	publisher = {TalkBank},
	author = {{Linguistic Data Consortium}},
	year = {2013},
}

@misc{graff_switchboard-2_1998-1,
	title = {Switchboard-2 {Phase} {I} ({LDC98S75})},
	url = {https://catalog.ldc.upenn.edu/LDC98S75},
	publisher = {Linguistic Data Consortium},
	author = {Graff, David and Canavan, Alexandra and Zipperlen, George},
	year = {1998},
	doi = {10.35111/c7th-nf28},
	note = {Place: Philadelphia
Published: Web Download},
}

@inproceedings{paul_design_1992,
	title = {The {Design} for the {Wall} {Street} {Journal}-based {CSR} {Corpus}},
	url = {https://aclanthology.org/H92-1073/},
	booktitle = {Speech and {Natural} {Language}: {Proceedings} of a {Workshop} {Held} at {Harriman}, {New} {York}, {February} 23-26, 1992},
	author = {Paul, Douglas B. and Baker, Janet M.},
	year = {1992},
}

@misc{canavan_callhome_1997,
	title = {{CALLHOME} {American} {English} {Speech} ({LDC97S42})},
	url = {https://catalog.ldc.upenn.edu/LDC97S42},
	publisher = {Linguistic Data Consortium},
	author = {Canavan, Alexandra and Graff, David and Zipperlen, George},
	year = {1997},
	doi = {10.35111/exq3-x930},
	note = {ISBN: 1-58563-111-6
Place: Philadelphia
Published: Web Download},
}

@inproceedings{bender2021dangers,
  title={On the dangers of stochastic parrots: Can language models be too big?},
  author={Bender, Emily M and Gebru, Timnit and McMillan-Major, Angelina and Shmitchell, Shmargaret},
  booktitle={Proceedings of the 2021 ACM conference on fairness, accountability, and transparency},
  pages={610--623},
  year={2021}
}

@article{zhao2018l2,
  title={L2-arctic: A non-native english speech corpus},
  author={Zhao, Guanlong and Chukharev-Hudilainen, Evgeny and Sonsaat, Sinem and Silpachai, Alif and Lucic, Ivana and Gutierrez-Osuna, Ricardo and Levis, John},
  year={2018}
}

@book{efron1994introduction,
  title={An introduction to the bootstrap},
  author={Efron, Bradley and Tibshirani, Robert J},
  year={1994},
  publisher={Chapman and Hall/CRC}
}

@inproceedings{rousseau-etal-2012-ted,
    title = "{TED}-{LIUM}: an Automatic Speech Recognition dedicated corpus",
    author = "Rousseau, Anthony  and
      Del{\'e}glise, Paul  and
      Est{\`e}ve, Yannick",
    editor = "Calzolari, Nicoletta  and
      Choukri, Khalid  and
      Declerck, Thierry  and
      Do{\u{g}}an, Mehmet U{\u{g}}ur  and
      Maegaard, Bente  and
      Mariani, Joseph  and
      Moreno, Asuncion  and
      Odijk, Jan  and
      Piperidis, Stelios",
    booktitle = "Proceedings of {LREC}",
    month = may,
    year = "2012",
    address = "Istanbul, Turkey",
    publisher = "European Language Resources Association (ELRA)",
    url = "https://aclanthology.org/L12-1405/",
    pages = "125--129"
}

@inproceedings{gururangan-etal-2020-dont,
    title = "Don{'}t Stop Pretraining: Adapt Language Models to Domains and Tasks",
    author = "Gururangan, Suchin  and
      Marasovi{\'c}, Ana  and
      Swayamdipta, Swabha  and
      Lo, Kyle  and
      Beltagy, Iz  and
      Downey, Doug  and
      Smith, Noah A.",
    editor = "Jurafsky, Dan  and
      Chai, Joyce  and
      Schluter, Natalie  and
      Tetreault, Joel",
    booktitle = "Proceedings of the 58th Annual Meeting of the Association for Computational Linguistics",
    month = jul,
    year = "2020",
    address = "Online",
    publisher = "Association for Computational Linguistics",
    url = "https://aclanthology.org/2020.acl-main.740/",
    doi = "10.18653/v1/2020.acl-main.740",
    pages = "8342--8360"
}

@article{chan2021speechstew,
  title={Speechstew: Simply mix all available speech recognition data to train one large neural network},
  author={Chan, William and Park, Daniel and Lee, Chris and Zhang, Yu and Le, Quoc and Norouzi, Mohammad},
  journal={arXiv preprint arXiv:2104.02133},
  year={2021}
}

@inproceedings{meyer-etal-2020-artie,
    title = "Artie Bias Corpus: An Open Dataset for Detecting Demographic Bias in Speech Applications",
    author = "Meyer, Josh  and
      Rauchenstein, Lindy  and
      Eisenberg, Joshua D.  and
      Howell, Nicholas",
    editor = "Calzolari, Nicoletta  and
      B{\'e}chet, Fr{\'e}d{\'e}ric  and
      Blache, Philippe  and
      Choukri, Khalid  and
      Cieri, Christopher  and
      Declerck, Thierry  and
      Goggi, Sara  and
      Isahara, Hitoshi  and
      Maegaard, Bente  and
      Mariani, Joseph  and
      Mazo, H{\'e}l{\`e}ne  and
      Moreno, Asuncion  and
      Odijk, Jan  and
      Piperidis, Stelios",
    booktitle = "Proceedings of the Twelfth Language Resources and Evaluation Conference",
    month = may,
    year = "2020",
    address = "Marseille, France",
    publisher = "European Language Resources Association",
    url = "https://aclanthology.org/2020.lrec-1.796/",
    pages = "6462--6468",
    language = "eng",
    ISBN = "979-10-95546-34-4"
}

@article{del2021earnings,
  title={Earnings-21: A practical benchmark for asr in the wild},
  author={Del Rio, Miguel and Delworth, Natalie and Westerman, Ryan and Huang, Michelle and Bhandari, Nishchal and Palakapilly, Joseph and McNamara, Quinten and Dong, Joshua and Zelasko, Piotr and Jett{\'e}, Miguel},
  journal={arXiv preprint arXiv:2104.11348},
  year={2021}
}

@article{o2021spgispeech,
  title={Spgispeech: 5,000 hours of transcribed financial audio for fully formatted end-to-end speech recognition},
  author={O'Neill, Patrick K and Lavrukhin, Vitaly and Majumdar, Somshubra and Noroozi, Vahid and Zhang, Yuekai and Kuchaiev, Oleksii and Balam, Jagadeesh and Dovzhenko, Yuliya and Freyberg, Keenan and Shulman, Michael D and others},
  journal={arXiv preprint arXiv:2104.02014},
  year={2021}
}

@misc{facebookresearch_voxpopuli,
  title        = {{facebookresearch/voxpopuli}: A large-scale multilingual speech corpus for representation learning, semi-supervised learning and interpretation},
  author       = {{Facebook Research}},
  year         = {2021},
  howpublished = {\url{https://github.com/facebookresearch/voxpopuli}},
  note         = {GitHub repository (archived)},
}

@article{kendall2018corpus,
  title={The corpus of regional african american language},
  author={Kendall, Tyler and Farrington, Charlie},
  journal={Version},
  volume={6},
  pages={1},
  year={2018}
}

@article{ott2022mapping,
  title={Mapping global dynamics of benchmark creation and saturation in artificial intelligence},
  author={Ott, Simon and Barbosa-Silva, Adriano and Blagec, Kathrin and Brauner, Jan and Samwald, Matthias},
  journal={Nature Communications},
  volume={13},
  number={1},
  pages={6793},
  year={2022},
  publisher={Nature Publishing Group UK London}
}

@article{sagawa2019distributionally,
  title={Distributionally robust neural networks for group shifts: On the importance of regularization for worst-case generalization},
  author={Sagawa, Shiori and Koh, Pang Wei and Hashimoto, Tatsunori B and Liang, Percy},
  journal={arXiv preprint arXiv:1911.08731},
  year={2019}
}

@article{barrault2023seamless,
  title={Seamless: Multilingual Expressive and Streaming Speech Translation},
  author={Barrault, Lo{\"\i}c and Chung, Yu-An and Meglioli, Mariano Coria and Dale, David and Dong, Ning and Duppenthaler, Mark and Duquenne, Paul-Ambroise and Ellis, Brian and Elsahar, Hady and Haaheim, Justin and others},
  journal={arXiv preprint arXiv:2312.05187},
  year={2023}
}

@article{kaplan2020scaling,
  title={Scaling laws for neural language models},
  author={Kaplan, Jared and McCandlish, Sam and Henighan, Tom and Brown, Tom B and Chess, Benjamin and Child, Rewon and Gray, Scott and Radford, Alec and Wu, Jeffrey and Amodei, Dario},
  journal={arXiv preprint arXiv:2001.08361},
  year={2020}
}

@article{recht2025actuary,
  title={The Actuary's Final Word on Algorithmic Decision Making},
  author={Recht, Benjamin},
  journal={arXiv preprint arXiv:2509.04546},
  year={2025}
}

@inproceedings{demirsahin-etal-2020-open,
    title = {{Open-source Multi-speaker Corpora of the English Accents in the British Isles}},
    author = {Demirsahin, Isin and Kjartansson, Oddur and Gutkin, Alexander and Rivera, Clara},
    booktitle = {Proceedings of {LREC}},
    month = may,
    year = {2020},
    pages = {6532--6541},
    address = {Marseille, France},
    publisher = {European Language Resources Association (ELRA)},
    url = {https://www.aclweb.org/anthology/2020.lrec-1.804},
    ISBN = {979-10-95546-34-4},
  }

@article{nilsson1994development,
  title={Development of the Hearing in Noise Test for the measurement of speech reception thresholds in quiet and in noise},
  author={Nilsson, Michael and Soli, Sigfrid D and Sullivan, Jean A},
  journal={The journal of the acoustical society of America},
  volume={95},
  number={2},
  pages={1085--1099},
  year={1994},
  publisher={Acoustical Society of America}
}

@misc{english_broadcast_news_hub4,
  author = {David Graff and John S. Garofolo and Jonathan G. Fiscus and William Fisher and David Pallett},
  title = {{1996 English Broadcast News Speech (HUB4)}},
  howpublished = {LDC97S44. Web Download. Philadelphia: Linguistic Data Consortium},
  year = {1997},
  doi = {10.35111/hvkc-4n95},
  url = {https://catalog.ldc.upenn.edu/LDC97S44}
}

\clearpage
\appendix
\subsection{Benchmark Details}
The specific benchmark datasets used by each model in our analysis are listed in Table \ref{tab:model-selection-details}. 

\begin{table*}[t]
\centering
    \caption{Models evaluated in our diachronic analysis. Parentheses indicate additional versions from the same model family. Datasets that explicitly evaluate accent variety within English are underlined.}
\begin{adjustbox}{max width=\textwidth}
\begin{tabular}{lcll>{\raggedright\arraybackslash}p{8cm}}
        \toprule
        \textbf{Model} & \textbf{Release date} & \textbf{Organization} & \textbf{Paper} & \textbf{Evaluation Datasets (ASR)} \\
        \midrule
          Deepspeech ($m_1$) & 2021-10 & Mozilla & \cite{amodei_deep_2016} & WSJ\cite{paul_design_1992},Librispeech\cite{librispeech_datasets},\underline{VoxForge}\cite{voxforge_eval_dataset},CHiME\cite{CHiME_eval_dataset}, \underline{CommonVoice-EN}\cite{commonvoice_datasets},Multilingual Librispeech\cite{pratap2020mls}\\
         WavLM ($m_2$) & 2021-12 & Microsoft & \cite{chen_wavlm_2021} & Librispeech\cite{librispeech_datasets}\\
         whisper ($m_3, (m_4, m_8)$) & 2022-09 & OpenAI & \cite{radford_robust_2022} & Librispeech\cite{librispeech_datasets},WSJ\cite{paul_design_1992},Switchboard\cite{graff_switchboard-2_1998-1},CORAAL\cite{kendall2018corpus}, TedLium\cite{rousseau-etal-2012-ted},CallHome\cite{canavan_callhome_1997},\underline{\text{VoxPopuli En}}\cite{wang-etal-2021-voxpopuli}, Fleurs\cite{conneau_fleurs_2023},\underline{AMI Meeting corpus}\cite{ami_meeting_eval_dataset},CommonVoice\cite{commonvoice_datasets}, CHiME\cite{CHiME_eval_dataset},\underline{Artie Bias Corpus}\cite{meyer-etal-2020-artie},Earnings21\cite{del2021earnings},\underline{Earnings22}\cite{earnings22_eval_dataset}\\
         MMS ($m_5$) & 2023-05 & Meta & \cite{pratap_scaling_2023} & Fleurs\cite{conneau_fleurs_2023},CommonVoice\cite{commonvoice_datasets},VoxPopuli\cite{wang-etal-2021-voxpopuli},Multilingual Librispeech\cite{pratap2020mls},MMS-lab (bible)\cite{pratap_scaling_2023}\\
         OWSM ($m_6 (m_7, m_{10}, m_{12})$) & 2023-09 & CMU & \cite{peng_reproducing_2023} & \underline{CommonVoice-EN}\cite{commonvoice_datasets},Fleurs\cite{conneau_fleurs_2023},Librispeech\cite{librispeech_datasets}, TedLium\cite{rousseau-etal-2012-ted},VoxPopuli\cite{wang-etal-2021-voxpopuli},WSJ\cite{paul_design_1992},Switchboard\cite{graff_switchboard-2_1998-1}, Multilingual Librispeech\cite{pratap2020mls}\\
         SeamlessM4T ($m_9$ ) & 2023-12 &  Meta & \cite{barrault2023seamless} & Fleurs \cite{conneau_fleurs_2023} \\
         Qwen-Audio ($m_{11} (m_{13})$) & 2024-08 & Alibaba & \cite{chu_qwen2-audio_2024} & Librispeech\cite{librispeech_datasets},\underline{CommonVoice-EN}\cite{commonvoice_datasets},Fleurs \cite{conneau_fleurs_2023}\\
         Canary-Qwen ($m_{14}$ ) &  2025-07 & NVIDIA & \cite{puvvada_less_2024} & \underline{AMI Meeting corpus}\cite{ami_meeting_eval_dataset},\underline{Earnings22}\cite{earnings22_eval_dataset},Gigaspeech\cite{gigaspeech_dataset}, \underline{SPGIspeech}\cite{o2021spgispeech},Tedlium\cite{rousseau-etal-2012-ted},Voxpopuli\cite{wang-etal-2021-voxpopuli},MUSAN\cite{musan_eval_dataset}, \text{CasualConversations Dataset}\cite{casual_conv_eval_dataset},\text{Multilingual Librispeech}\cite{pratap2020mls}\\
         Chirp-3 ($m_{15}$) & 2025-10 & Google & \cite{zhang_google_2023} & \underline{AMI Meeting Corpus}\cite{ami_meeting_eval_dataset},\text{Broadcast News}\cite{english_broadcast_news_hub4},Common Voice\cite{commonvoice_datasets},LibriSpeech\cite{librispeech_datasets},Switchboard\cite{graff_switchboard-2_1998-1},Tedlium\cite{rousseau-etal-2012-ted},WSJ, Fleurs\cite{conneau_fleurs_2023},Youtube captions\\ 
         Omnilingual ($m_{16}$) & 2025-11 & Meta & \cite{omnilingual_asr_team_omnilingual_2025} & \text{MMS-Lab (bible)}\cite{pratap_scaling_2023}, Fleurs\cite{conneau_fleurs_2023},MLS\cite{pratap2020mls},CommonVoice\cite{commonvoice_datasets}, AllASR\cite{omnilingual_asr_team_omnilingual_2025}\\
         \bottomrule
    \end{tabular}
    \end{adjustbox}
    \label{tab:model-selection-details}
\end{table*}

\begin{table*}[t]
\centering
\caption{Datasets that are explictly concerned with minority accent representation.}
\begin{adjustbox}{max width=\textwidth}
\begin{tabular}{l p{\dimexpr\textwidth-3cm\relax}}
\toprule
     VoxForge & ``Our source for accented speech is the publicly available VoxForge (http://www.voxforge.org)
dataset, which has clean speech read from speakers with many different accents.'' \cite{amodei_deep_2016}  \\
   Artie & ``The Artie Bias Corpus contains information on 3 gender
classes, 17 English accents, and 8 age ranges.'' \cite{meyer-etal-2020-artie}\\
    VoxPopuli-En & ``VoxPopuli provides 29 hours of transcribed speech data of non-native English intended for research in ASR for accented speech'' \cite{facebookresearch_voxpopuli} \\
    Earnings22 & ``Our attention focused on aggregation of
accented public English-language earnings calls from global
companies.'' \cite{earnings22_eval_dataset} \\
    SPGISpeech & ``There are roughly 50,000 speakers in SPGISpeech, drawn
from corporate officers and analysts appearing in English earnings calls and spanning a broad cross-section of L1 and L2 accents'' \cite{o2021spgispeech}\\
    CommonVoice-EN & ``The 2021 release of Common Voice English (7.0) contains 2,015 hours of (validated) speech
submitted by over 75,000 speakers some of whom opted to provide some information about their gender and accent'' \cite{markl-2022-mind}\\
    AMI Meeting Corpus & ``The meetings were recorded in English using three different rooms with different acoustic properties, and include mostly non-native speakers'' \cite{ami_meeting_eval_dataset}\\
\bottomrule
\end{tabular}
\end{adjustbox}
\label{tab:dataset-minority-representation-discourse}
\end{table*}
In Table \ref{tab:dataset-minority-representation-discourse}, we list the subset of benchmarks that are explicitly concerned with multi-accent representation, as demonstrated by pull quotes from the respective papers describing the benchmark.

\subsection{Dataset Descriptions} 

\paragraph{Speech Accent Archive} Speech Accent Archive \cite{weinberger2015speech} is an English speech corpus containing both non-standard and standard speakers reading the same passage: ``\textit{Please call Stella.  Ask her to bring these things with her from the store:  Six spoons of fresh snow peas, five thick slabs of blue cheese, and maybe a snack for her brother Bob.  We also need a small plastic snake and a big toy frog for the kids.  She can scoop these things into three red bags, and we will go meet her Wednesday at the train station.}''

It has previously been used for assessing performance disparities in ASR systems across different Global English varieties \cite{dichristofano2022global,chan2022training,graham_evaluating_2024,weinberger2015speech}. 
The dataset controls for the noise level and the distance to the microphone. Each subject is given one minute to read over the passage and ask questions about unfamiliar words before recording. However, this was done by many different researchers, so the quality and procedure does differ between samples. As of writing, the corpus contains 3038 samples across 392 standard languages. The corpus contains a smaller subset of high quality expert-annotated samples. This constitutes approximately 10 hours of speech from 1238 speakers across 831 birthplaces and 190 standard languages. Of these speakers, 181 are listed as standard English speakers. We only kept accents where $20+$ speakers had contributed a recording. Each sample has median duration of 26 seconds (8 second std.), varying by speaking rate.

\paragraph{L2 ARCTIC}
L2 ARCTIC \cite{zhao2018l2} contains about 1 hour of scripted English from each of 24 non-standard speakers, utilizing source texts originally derived from Project Gutenberg \cite{kominek_cmu_database}. The speakers are evenly divided, 2 male and 2 female, between standard Arabic, Hindi, Vietnamese, Spanish, Korean, and Mandarin. There is also an unscripted subset, the ``suitcase" subset from 22 of the speakers where they were prompted with the suitcase story \cite{derwing_suitcase_l1_l2} and asked to explain what was happening in the pictures. About 1 hour of the scripted subset is expert-annotated (3,599 samples, 150 utterances for each speaker). The entire suitcase subset is expert-annotated. Additionally, we record 1 sample of L1 North American English speech by one of the authors using the suitcase prompt, given that L2 Arctic does not contain an unscripted sample spoken by a standard English speaker.

\paragraph{OpenSLR83}
OpenSLR83 is a dataset of transcribed speech featuring English speakers from various dialectal regions across the UK and Ireland. The original dataset comprises 17,877 utterances across nine dialect-gender combinations: Irish English, Midlands English, Northern English, Scottish English, Southern English, and Welsh English.

\paragraph{ALLSSTAR} ALLSSTAR Corpus \cite{bradlow2010allstar} is a multilingual speech dataset containing paired recordings from each participant in both their first language (L1), and in English as a second language (L2). The dataset features 102 L2 English speakers representing 21 standard language backgrounds, along with 26 standard English speakers.

The corpus comprises both read and spontaneous speech (e.g., picture descriptions and question–answer sessions); we focused on the read-speech portion for this dataset. The read materials include short sentences and paragraphs, with each speaker producing 120 unique sentences and 3 unique paragraphs. Sentence recordings were collected in two sessions, each captured as a single continuous audio file. The paragraph topics were: Little Prince (LPP), Declaration of Human Rights (DHR), and The North Wind and the Sun (NWS). In total, the dataset includes 129 recordings from L1 English speakers and 570 from L2 English speakers. The final analyzed subset comprised 10,753 short sentences (2,862 L1; 7891 L2), 2,548 LPP recordings (727 L1; 1,821 L2), and 1,869 DHR paragraphs (514 L1; 1,355 L2). Recording duration varied by material type, with short sentences averaging 1.68 seconds, LPP averaging 4.16 seconds, and DHR paragraphs averaging 6.04 seconds.

\subsection{Preprocessing Details} \label{app:filtering}
Here we provide additional details for filtering out some samples from the evaluation datasets we applied in our analysis.

\paragraph{Speech Accent Archive.} Each recording is published as a single wave file. The expert annotated subset comes with phonetic and text transcriptions in a Praat TextGrid file. We keep only the expert annotated subset. We remove untranscribed sections of the audio labeled as ``chatter'', ``self-introduction'', ``self-talk'', ``experimenter comments'', ``background speech'', ``noise'', ``bird song'', ``cough'' and any variant mispellings thereof. Furthermore, we standardize numbers to text form (``2'' $\to$ ``two''), fix typos, and clip audio into consistent duration segments of 50 phonemes each.

\paragraph{L2-Arctic.} Each utterance is published as a single wave file with a corresponding TextGrid file containing phonetic and text transcriptions. Except for the expert-annotated subset, these are machine generated. We keep only the high quality, expert-annotated subsets. In this subset, each scripted utterance is about 3.6 seconds of speech. We clip the suitcase utterances to be approximately the same length. As part of the clipping process, we remove audio that hasn't been annotated, e.g., the interviewer speaking or the subject mumbling.

\paragraph{OpenSLR83.} We remove utterances containing fewer than 4 words. To maintain balanced representation across dialects and genders, we select 100 samples per dialect with equal gender distribution (50 female, 50 male samples per dialect). For Irish English, which contains only male speakers, we select 50 samples. This yields a balanced subset for analysis.

\paragraph{ALLSTAR.} 
Each recording is accompanied by Praat TextGrid annotations at the sentence, word, and phoneme levels. For the present analysis, full-session audio files were segmented into individual sentence-level utterances based on these annotations, and NWS paragraph recordings were excluded due to the absence of utterance-level time alignments. Instances in which the recordings did not correspond to the reference text were automatically filtered out by removing utterances for which all ASR models produced a word error rate greater than 100.

\subsection{Accent and Demographic Information}
\begin{table*}[htbp]
\centering
\small
\caption{Summary of evaluation datasets. Duration is given in seconds per sample as the interquartile range (IQR). All datasets contain scripted speech except L2-Arctic Spontaneous.}
\renewcommand{\arraystretch}{1.3}
\begin{tabular}{p{2cm} p{1.8cm} p{1.5cm} p{3cm} p{4.5cm}}
\toprule
\textbf{Dataset} & \textbf{Duration (s, IQR)} & \textbf{Samples} & \textbf{Accents} & \textbf{Speaker Demographics} \\
\midrule
Speech Accent Archive  
  & 5.4--7.2 & 1,238 
  & nld, spa, ita, eng, fra, kor, tur, ara, por, deu, zho, rus 
  & Mean age 33.31 (SD 13.57); started English at 10.53 (SD 8.49); 55.7\% male. \\

L2-Arctic 
  & 2.8--4.4 & 3,599 
  & vie, cmn, hin, spa, ara, kor 
  & Iowa State University students; age 22--43, mean 29 (SD 6.9). \\

L2-Arctic Spontaneous 
  & 6.2--7.3 & 22 
  & vie, cmn, hin, spa, ara, kor 
  & Same as L2-Arctic. \\

OpenSLR83 
  & 4.6--7.7 & 17,877 
  & Irish, Midlands, Northern, Scottish, Southern, Welsh 
  & Volunteers 21+; recruited by Google London and Cardiff University. \\

ALLSTAR  
  & 1.5--2.4 & 42,398 
  & pbr, cct, kor, guj, vie, ctw, run, hin, fra, rus, gis, csp, gre, ind, spa, heb, tur, ger, far, shs, cmn, jpn 
  & Northwestern University students; age 18--41, mean 23.52. \\
\bottomrule
\end{tabular}
\label{tab:dataset-details}
\end{table*}
See Table \ref{tab:dataset-details}.

\subsection{System Degradation Examples}
\begin{table*}[t]
\centering
\small
\caption{Illustration of degradation from earlier model $m_i$ to later model $m_j$ across all of our evaluation datasets.}
\begin{tabular}{p{2.5cm}p{1.1cm}p{4.6cm}p{4.2cm}p{3.3cm}}
\textbf{Dataset (Accent)} &
\hspace{-8px}$\mathbf{\Bar{d}_{(m_i\to m_j)}}$ &
\textbf{Prediction ($m_i$)} &
\textbf{Prediction ($m_j$)} &
\textbf{Transcript} \\
\toprule
L2-Arctic \qquad (Vietnamese) &
12.5\% &
Whisper Large v2 (2022-12): Without a doubt, some of them have dinner engagements. &
MMS 1B All (2023-05): wit dou a doubt some of them hapdiner engagements &
without a doubt some of them have dinner engagements \\
\midrule
L2-Arctic \qquad Spontaneous \qquad (Korean) &
15.4\% &
Canary (2025-07): Okay, um I can see skyscrapers tall buildings and I think &
Omnilingual 7B (2025-11): i can see skyscripers tall buildings and i think &
ok uh i can see skyscrapers tall buildings and i think \\
\midrule
OpenSLR-83 (Irish) &
5.7\% &
Google Chirp 3 (2025-10): Your app facilitates or promotes content that includes gratuitous violence or dangerous activities. &
Omnilingual 7B (2025-11): your off facilitates or promotes content that includes gratuitous flylence or dangerous activities &
Your app facilitates or promotes content that includes gratuitous violence or dangerous activities \\
\midrule
SAA \qquad \qquad (Korean) &
18.3\% &
Whisper Large v2 (2022-12): Please call Stella. Ask her to bring these things with her from the store. &
MMS 1B All (2023-05): please costella escr to bring these things with her from the store &
please call stella Ask her to bring these things with her from the store. \\
\midrule
ALLSTAR-HT1 (Nkore) &
6.7\% &
Whisper Large v1 (2022-09): the salt shaker is empty &
Canary (2025-07): the sword sheikah is empty &
the salt shaker is empty \\
\midrule
ALLSTAR-HT2 (Nkore) &
3.1\% &
Whisper Large v1 (2022-09): he is washing his face with soap &
Omnilingual (2025-11): he is washing his face with sore &
he is washing his face with soap \\
\midrule
ALLSTAR-LPP (Nkore) &
10.1\% &
Whisper Large v2 (2022-12): draw me a sheep &
Canary (2025-07): turn me a say &
draw me a sheep \\
\midrule
ALLSTAR-DHR (Cantonese) &
6.1\% &
Whisper Large v3 (2023-11): everyone has the right to a nationality &
OWSM 3.1 (2024-01): every one has to wai t t t snr t &
everyone has the right to a nationality \\
\bottomrule
\end{tabular}
\label{tab:degradation-example-table}
\end{table*}

In Table~\ref{tab:degradation-example-table}, we list examples of other pairs of models, 
one pair for each dataset 
we evaluated on the minority accent we highlighted in Figure~\ref{fig:regression-plots}. 
For the selected pair of models $m_i$ and $m_j$, 
we compute the average degradation across the entire dataset, 
listed in the second column of Table~\ref{tab:degradation-example-table}, and select a 
random example from the top quartile of the degradation distribution across the dataset.

\end{document}